\documentclass[smallextended]{svjour3}      
\usepackage{lineno,hyperref}
\modulolinenumbers[1]
\usepackage{amssymb}
\usepackage{amsmath}
\usepackage{pdflscape}
\usepackage{array,multirow}
\usepackage{caption}
\usepackage{rotating}
\usepackage{graphicx}
\usepackage{subfig}
\usepackage{tabularx}
\usepackage{multirow}
\usepackage{lmodern}
\usepackage{lscape}
\usepackage{changepage}

\begin{document}

\title{ A  $\nu$- support vector quantile regression model with automatic accuracy control}

	\titlerunning{~~~~}      

\author{ Pritam Anand \and Reshma Rastogi (nee Khemchandani) \and Suresh Chandra. 
}

\institute{   Pritam Anand \\
	Research Scholar, \at
	Faculty of Mathematics and Computer Science,\\
	South Asian University, New Delhi-110021.\\
	\email{ltpritamanand@gmail.com ~.} 
	\and 
	R. Rastogi (nee Khemchandani)   \\
	Assistant Professor,             \at
	Faculty of Mathematics and Computer Science,\\
	South Asian University, New Delhi-110021. \\
	\email{reshma.khemchandani@sau.ac.in .}  \\
	Tel No. 01124195147        
	\and
	Suresh Chandra\at
	Ex-Faculty, Department of Mathematics\\
	Indian Institute of Technology Delhi.
	New Delhi-110016.\\
	\email{chandras@maths.iitd.ac.in}                            
}

\date{Received: date / Accepted: date}

\maketitle
\begin{abstract}
This paper proposes a novel '$\nu$-support vector quantile regression' ($\nu$-SVQR) model for the   quantile estimation. It can facilitate the automatic control over accuracy by creating a suitable asymmetric $\epsilon$-insensitive zone according to the variance present in data. The proposed $\nu$-SVQR model uses the $\nu$ fraction of training data points for the estimation of the quantiles. In the $\nu$-SVQR model, training points asymptotically appear above and below of the asymmetric $\epsilon$-insensitive tube in the ratio of $1-\tau$ and $\tau$. Further, there are other interesting properties of the proposed $\nu$-SVQR model, which we have briefly described in this paper.  These properties have been empirically verified using the artificial and real world dataset also.
\end{abstract}

\keywords{
\texttt { Quantile Regression},  pinball loss function , Support Vector Machine, $\epsilon$-insensitive loss function. 
}

\section{Introduction}
Given the training set $T= \{ (x_i,y_i): x_i \in \mathbb{R}^n, y_i \in \mathbb{R},~ i=1,2...,l~ \}$ and $\tau \in [0,1]$, the problem of quantile regression is to estimate a real valued function $f_\tau(x)$ such that a proportion $\tau$ of $y/x$ will be lying below of the estimate $f_\tau(x)$. For $\tau=0.5$, the problem is equivalent to median estimation. The estimation of $f_\tau(x)$ is difficult but, more informative than estimation of only mean regression $f(x)$. The estimation of $f_\tau(x)$ for different values of $\tau$ can briefly describe the different characteristics of the conditional distribution of $y/x$. In many real world problems, the estimation of mean regression $f(x)$ is not required or enough, rather they require the estimation of quantile  $f_\tau(x)$.  

 The study of quantile regression problem has initially been started in 1978 by Koenkar and Bassett\cite{quantile1}. Later, it has been briefly discussed and described by Koenker in his book (Koenker, \cite{quantile2}). Koenkar and Bassett \cite{quantile1} proposed the pinball loss function for the estimation of the quantile function  $f_{\tau}(x)$. For a given quantile $\tau \in (0,1)$, the pinball loss function was an asymmetric loss function suitable for quantile estimation. It was given by
 \begin{equation}
 P_{\tau}(u) ~=~ \begin{cases}
 \tau u ~~~~~~~~~~\mbox{if}~~ u \geq 0, \\
 (\tau-1)u~~~ \mbox{otherwise}.
 \end{cases}
 \label{pinballloss}
 \end{equation}

 Support Vector Regression (SVR) models (Vapnik et al.,\cite{svr1})(Drucker et al.,\cite{svr2}),(Gunn, \cite{GUNNSVM}) are one of the most popular regression model which can estimate the mean regression function $f(x)$ efficiently. SVR model commonly solves a Convex Program which guarantees the global optimal solution. These models have been widely used in solving real world problems of diverse domain. 
 
 Takeuchi et al \cite{quantile3} initiated the study of the quantile regression problem in a non-parametric framework on the line of SVR models. They have proposed Support Vector Quantile Regression (SVQR) model in which they have minimized the pinball loss function in SVR type optimization problem for estimation of the quantile function $f_\tau(x)$. The obtained solution of SVQR model is not sparse as every training data points are allowed to contribute in the empirical risk which is measured by the asymmetric pinball loss function.
 
 Researchers have attempted to extend the SVQR model on the line of $\epsilon$-SVR model for increasing its generalization ability as well as  obtaining the sparse solution. For this, they have attempted to propose the $\epsilon$-insensitive pinball loss functions to incorporate the concept of $\epsilon$-insensitive zone in the asymmetric pinball loss function. 
  
    At first, Takeuchi and  Furuhashi considered the $\epsilon$-insensitive pinball loss function for estimation of the non-crossing quantile in their work (Takeuchi and  Furuhashi, \cite{noncrossqsvr}). Further, Hu et al, had also considered the similar kind of $\epsilon$-insensitive pin ball loss function in their work (Hu et al, \cite{onlinesvqr}) for estimation of quantiles. However, the $\epsilon$-insensitive zone in these pinball loss function was symmetric. The use of the symmetric $\epsilon$ -insensitive zone in the asymmetric pinball loss function failed to perform well for estimation of quantiles. 
 
              Soek et al. have  first considered the asymmetric $\epsilon$-insensitive zone in the pinball loss function in their proposed e-sensitive pinball loss function (Soek et al., \cite{sparsequantile}). Later on, Park and Kim \cite{quantilerkhs} has also proposed a similar kind of loss function in their work (Park and Kim, \cite{quantilerkhs}). But problem of these pinball loss function was that they failed to provide a suitable $\epsilon$ -insensitive zone for every value of $\tau$. 
              
               Anand et al. have proposed an asymmetric $\epsilon$-insensitive pinball loss function in their work (Anand et al., \cite{anand}) which extends the concept of $\epsilon$-insensitive zone in the pinball loss function in true sense. The asymmetric $\epsilon$-insensitive pinball loss can obtain a suitable $\epsilon$-insensitive zone of fixed width for every values of $\tau$. The $\epsilon$ -insensitive zone was partitioned using $\tau$ value in the asymmetric $\epsilon$-insensitive pinball loss function. Using the asymmetric $\epsilon$-insensitive pinball loss function, they have proposed $\epsilon$ -SVQR model which can obtain better generalization ability than existing SVQR models and successfully brings the sparsity back in the SVQR model. 
               
               However, the $\epsilon$-SVQR model (Anand et al., \cite{anand}) requires a good choice of value of $\epsilon$ for obtaining the better prediction of quantiles. A bad choice of $\epsilon$ can distort the performance of the $\epsilon$-SVQR model (Anand et al., \cite{anand}). 
               
               This paper proposes an efficient SVQR model which appropriately trade-off the total width of the asymmetric $\epsilon$-insensitive zone in its optimization problem via the user defined parameter $\nu$. The proposed model has been termed with $\nu$-Support Vector Quantile Regression ($\nu$-SVQR) model. The $\nu$-SVQR model can adjust the  overall width of asymmetric $\epsilon$-insensitive zone such that at most $\nu$ fraction of training data points lie outside of it. This capability of $\nu$ -SVQR  enables it to automatically adjust the width of the $\epsilon$-insensitive zone according to the variance present in the data without adjusting any parameter. In the $\nu$-SVQR model, training points asymptotically appear above and below of the asymmetric $\epsilon$-insensitive tube in the ratio of $1-\tau$ and $\tau$. Further, there are other interesting asymptotic properties of $\nu$-SVQR model which we have briefly described in this paper. Several experiments on artificial as well as UCI datasets have been performed to empirically verify claims made in this paper. 
               
              The rest of this paper is organized as follows. Section-2 briefly describes the standard Support Vector Quantile Regression(SVQR) model\cite{quantile3} and $\epsilon$-Support Vector Quantile Regression ($\epsilon$-SVQR) model\cite{anand}. In Section-3, we  present our proposed $\nu$-Support Vector Quantile Regression ($\nu$-SVQR) model and its different properties. Section-4 contains the numerical results obtained by different nature of experiments carried on artificial as well as real world datasets to empirically verify the properties of proposed $\nu$-SVQR model.
              
  \section{Support Vector Quantile Regression models} 
    For the training set $T= \{ (x_i,y_i): x_i \in \mathbb{R}^n, y_i \in \mathbb{R},~ i=1,2...,l~ \}$ and the quantile $\tau \in (0,1)$ , the  SVQR model estimates the  function $f_{\tau}(x) = w^T\phi(x)+ b $ in the feature space for the estimation of the $\tau$th quantile, where $\phi:\mathbb{R}^n \rightarrow \mathcal{H}$ is a mapping from the input space to a higher dimensional feature space $\mathcal{H}$.
    \subsection{Standard Support Vector Quantile Regression model}   
    The standard Support Vector Quantile regression model minimizes
    \begin{eqnarray}
    \min_{w,b}~ \frac{1}{2}||w||^2 + C.\sum_{i=1}^{l}L_\tau({y_i-(w^Tx_i+b)}), \label{stan_svqr}
    \end{eqnarray}
     where $L_\tau(v)$ is the asymmetric pinball loss function which is given by
      \begin{equation}
        L_{\tau}(v) ~=~ \begin{cases}
        \tau v ~~~~~~~~~~\mbox{if}~~ v > 0, \\
        (\tau-1)v~~~ \mbox{otherwise}.
        \end{cases}
        \label{pinball_loss}
        \end{equation}
        Using the $l$-dimensional variables $\xi=(\xi_1,\xi_2,....,\xi_l)$ and $\xi^*=(\xi^*_1,\xi^*_2,....,\xi^*_l)$, the optimization problem (\ref{stan_svqr}) can be equivalently converted to following Quadratic Programming Problem (QPP)
        \begin{eqnarray}
        \min_{(w,b,\xi,\xi^*)}~~ \frac{1}{2}||w||^2 + C.\sum_{i=1}^{l}(\tau\xi_i+ (1-\tau)\xi_i^{*}) \nonumber \\
        & \hspace{-105mm}\mbox{subject to,}\nonumber\\
        & \hspace{-70mm}y_i- (w^T\phi(x_i)+b) \leq   \xi_i,  \nonumber\\
        & \hspace{-70mm}(w^T\phi(x_i)+b)-y_i \leq  \xi_i^{*} , \nonumber\\
        & \hspace{-60mm}\xi_i \geq 0,~~\xi_i^{*} \geq 0, ~~~ i =1,2,...l.
        \label{SVQR_primal}
        \end{eqnarray}
        Here $C \geq 0$ is a user defined parameter which is used to find a good trade-off between empirical risk and model complexity of estimator. The QPP (\ref{SVQR_primal}) of standard SVQR model can be easily solved  by solving its corresponding Wolfe dual problem. More detail about standard SVQR model can be found in (Takeuchi et al.,\cite{quantile3}).
    \subsection{$\epsilon$- Support Vector Quantile Regression model}    
    Anand et al.\cite{anand} have proposed an asymmetric $\epsilon$-insensitive pinball loss function which can obtain a suitable asymmetric $\epsilon$-insensitive zone for every values of $\tau$. The asymmetric $\epsilon$-insensitive pinball loss function is given by
    \begin{equation}
    L_{\tau}^{\epsilon}(u)= max(~-(1-\tau)(u+\tau\epsilon),~0~,~\tau(u-(1-\tau)\epsilon)~).
    \end{equation}
     It can be better understood in the following form
    \begin{equation}
    L_{\tau}^{\epsilon}(y_i,x_i,w,b)= 
    \begin{cases}
    -(1-\tau)(y_i-(w^Tx_i+b)+ \tau\epsilon), ~if~~y_i-(w^Tx_i+b) < -\tau\epsilon. \\
    0,         ~~~~~~~~~~~~~~~~~~~~~~~~if~~ -\tau\epsilon \leq y_i-(w^Tx_i+b)\leq (1-\tau)\epsilon. \\
    \tau(y_i-(w^Tx_i+b)-(1-\tau)\epsilon),~if ~y_i-(w^Tx_i+b) >(1-\tau)\epsilon.
    \end{cases}
    \end{equation}
    
     \begin{figure}
     	\centering
     	\subfloat[] {\includegraphics[width=2.0in,height=1.2in]{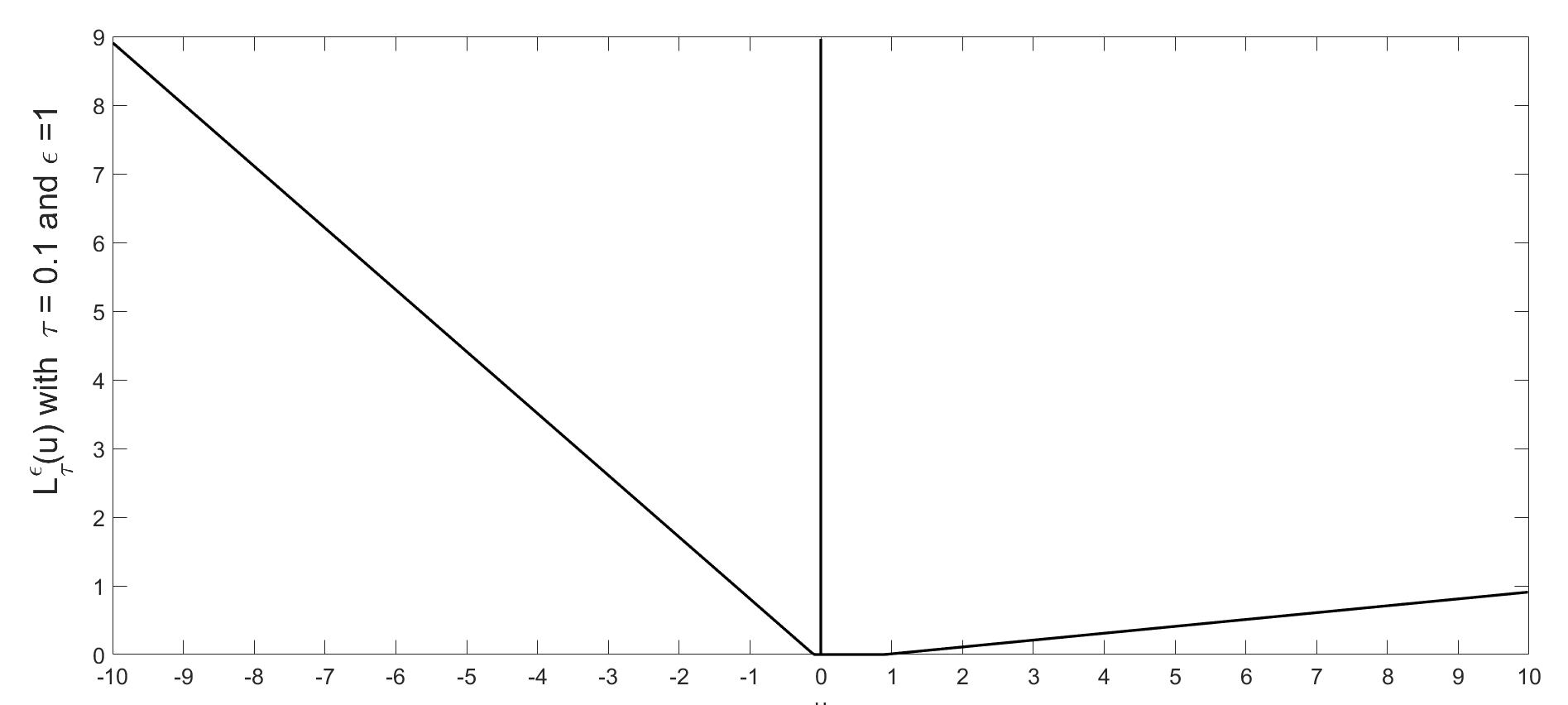}}
     	\subfloat[] {\includegraphics[width=0.48\linewidth]{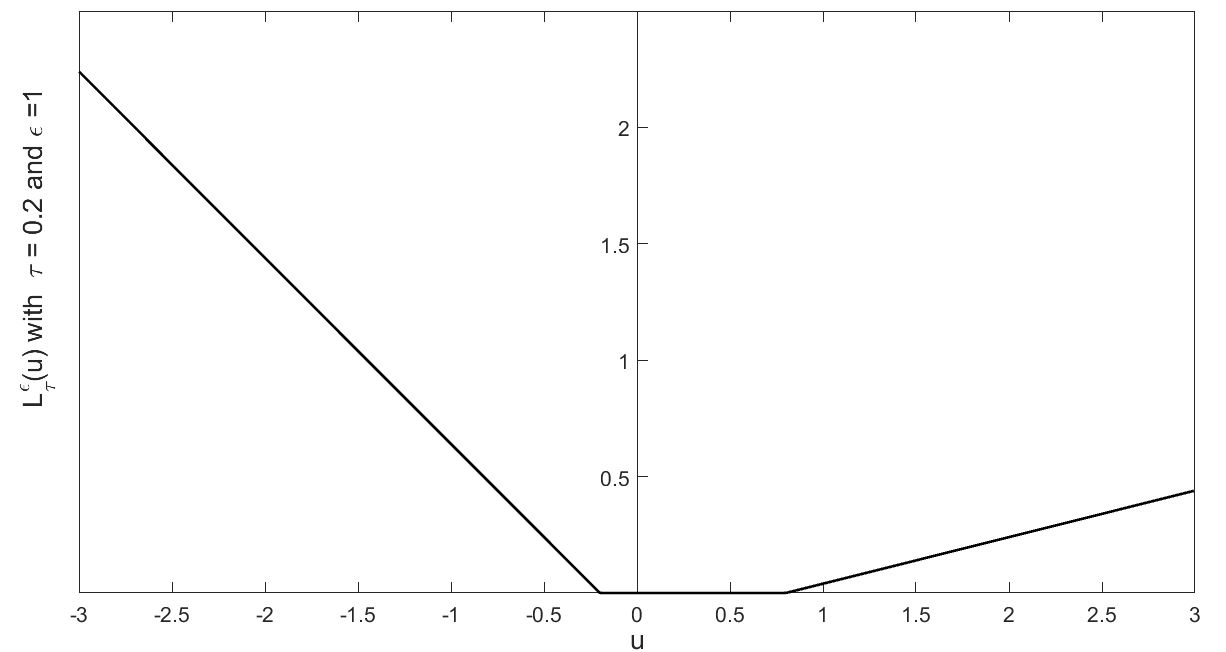}}\\
     	\subfloat[] {\includegraphics[width=2.0in,height=1.2in]{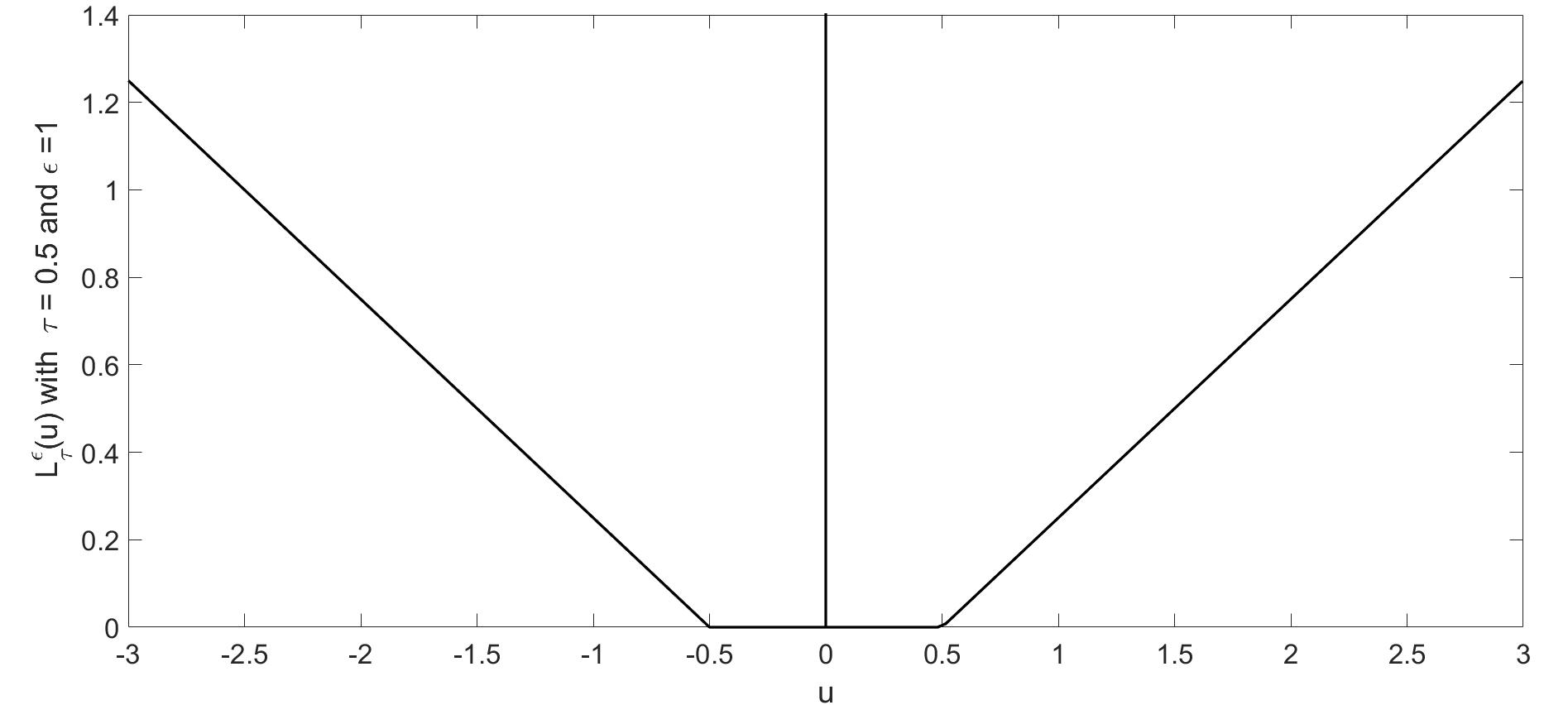}}
     	\subfloat[] {\includegraphics[width=0.48\linewidth]{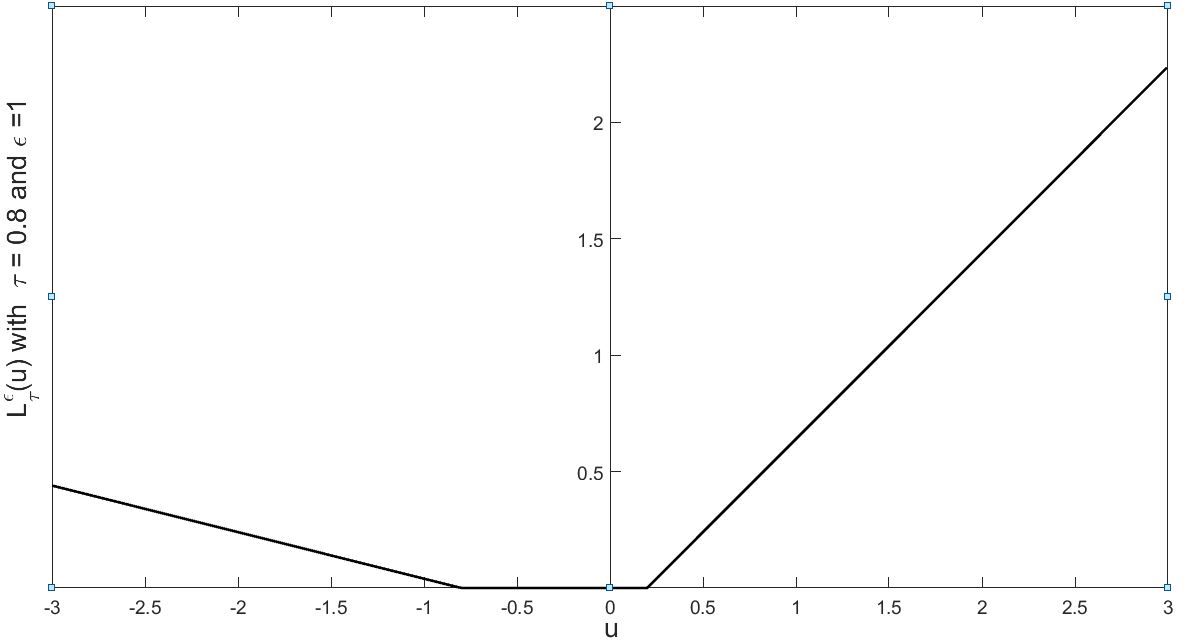}}\\
     	\caption{ The asymmetric $\epsilon$-pinball loss function for (a) $\tau =0.1$ (b) $\tau= 0.2$ (c) $\tau = 0.5$  and (d) $\tau = 0.8$  with fixed $\epsilon$=1.}
     	\label{ourloss}
     \end{figure}
     Figure (\ref{ourloss}) shows that the asymmetric $\epsilon$-insensitive pinball loss function can generate the suitable asymmetric $\epsilon$-insensitive zone for different values of $\tau$.
     
     The $\epsilon$-SVQR model minimizes
     \begin{eqnarray}
     \min_{(w,b)} ~\frac{1}{2}||w||^2 + C. \sum_{i=1}^{l}L_{\tau}^{\epsilon}(y_i,x_i,w,b)  \nonumber \\
     & \hspace{-60mm}   =  \min_{(w,b)}~ \frac{1}{2}||w||^2 + C. \sum_{i=1}^{l}max(-(1-\tau)(y_i-(w^Tx_i+b)+\tau\epsilon),  \nonumber \\ & \hspace {-05mm}0,
     \tau(y_i-(w^Tx_i+b)-(1-\tau)\epsilon)) 
     \end{eqnarray}
       which can be equivalently converted to following QPP
       \begin{eqnarray}
       \min_{(w,b,\xi,\xi^*)} \frac{1}{2}||w||^2 + C.\sum_{i=1}^{l}(\tau\xi_i+ (1-\tau)\xi_i^{*}) \nonumber \\
       & \hspace{-110mm}\mbox{subject to,}\nonumber\\
       & \hspace{-50mm}y_i- (w^T\phi(x_i)+b) \leq (1-\tau)\epsilon +  \xi_i,  \nonumber\\
       & \hspace{-60mm}(w^T\phi(x_i)+b)-y_i \leq   \tau\epsilon + \xi_i^{*} , \nonumber\\
       & \hspace{-60mm}\xi_i \geq 0,~~\xi_i^{*} \geq 0, ~~~ i =1,2,...l.
       \label{esvqrprimal}
       \end{eqnarray}
       In the $\epsilon$-SVQR model, $\epsilon \geq 0$ is the user defined parameter and a good value of $\epsilon$ is required beforehand for the efficient estimate of quantiles.
       For the solution of the $\epsilon$-SVQR primal problem (\ref{esvqrprimal}), we obtain its corresponding Wolfe dual problem as follows
       \begin{eqnarray}
       \min_{\alpha,\beta} \frac{1}{2}\sum_{i=1}^{l}\sum_{j=1}^{l}(\alpha_i- \beta_j)K(x_i,x_j)(\alpha_j-\beta_i) - \sum_{i=1}^{l}(\alpha_i-\beta_i)y_i + \sum_{i=1}^{l}((1-\tau)\epsilon\alpha_i+\tau\epsilon\beta_i) \nonumber\\
       & \hspace*{-240mm}\mbox{subject to,} \nonumber \\
       &\hspace{-190mm} \sum_{i=1}^{l}(\alpha_i-\beta_i)= 0, \nonumber \\
       & \hspace{-180mm}  0 \leq \alpha_i \leq C\tau,~~ i=~1,2,...l,       \nonumber \\
       & \hspace{-172mm} 0 \leq \beta_i \leq C(1-\tau), ~~i=~1,2,...l.  
       \label{esvqrdual}   
       \end{eqnarray}
       After obtaining the solution of the dual problem (\ref{esvqrdual}),  we can estimate $f_\tau(x) $, for any test data point $x \in \mathbb{R}^n$ using
       \begin{equation}
       f_\tau(x) = \sum_{i=1}^{l}(\alpha_i-\beta_i)K(x,x_i) + b.
       \end{equation}

  \section{Proposed $\nu$-Support Vector Quantile Regression model }
   The proposed $\nu$-SVQR model minimizes 
   \begin{eqnarray}
       \min_{(w,b,\epsilon, \xi,\xi^*)} \frac{1}{2}||w||^2 + C( \nu\tau(1-\tau)\epsilon + \frac{1}{l}\sum_{i=1}^{l}(\tau\xi_i+ (1-\tau)\xi_i^{*}) ) \nonumber \\
        & \hspace{-130mm}\mbox{subject to,}\nonumber\\
       & \hspace{-70mm}y_i- (w^T\phi(x_i)+b) \leq (1-\tau)\epsilon +  \xi_i,  \nonumber\\
       & \hspace{-80mm}(w^T\phi(x_i)+b)-y_i \leq   \tau\epsilon + \xi_i^{*} , \nonumber\\
        & \hspace{-70mm}\xi_i \geq 0,~~\xi_i^{*} \geq 0,~~\epsilon \geq 0, ~~~ i =1,2,...l,
        \label{esvqr_primal}
      \end{eqnarray}
  where $C \geq 0 $ and $\nu \geq 0$ are user defined parameters.

For solving the primal problem (\ref{esvqr_primal}) efficiently, we need to derive its  Wolfe dual problem. The Lagrangian function for the primal problem (\ref{esvqr_primal}) is obtained as
\begin{eqnarray}
 & \hspace{-130mm} L(w, b, \epsilon, \xi_i ,\xi_i^{*}, \alpha_i,\beta_i,\gamma_i,\lambda_i,\eta) = ~~~~ \frac{1}{2}||w||^2 + C( \nu\tau(1-\tau)\epsilon + \frac{1}{l}\sum_{i=1}^{l}(\tau\xi_i+ (1-\tau)\xi_i^{*}) )  \nonumber \\
+\sum_{i=1}^{l}\alpha_i(y_i- (w^T\phi(x_i)+b)- (1-\tau)\epsilon -  \xi_i)+ \sum_{i=1}^{l}\beta_i((w^T\phi(x_i)+b)-y_i - \tau\epsilon - \xi_i^{*})\nonumber \\   &\hspace{-220mm}-\sum_{i=1}^{l}\gamma_i\xi_i-
\sum_{i=1}^{l}\lambda _i\xi_i^{*} -\eta\epsilon
\end{eqnarray}
  We can now note the KKT conditions for (\ref{esvqr_primal}) as follows
 \begin{eqnarray}
  & \hspace{-90mm}\frac{\partial L}{\partial w} = w+ \sum_{i=1}^{l}(\beta_i-\alpha_i)\phi(x_i)=0 \implies w = \sum_{i=1}^{l}(\alpha_i-\beta_i)\phi(x_i)  \label{kkt1}\\
  & \hspace{-152mm}\frac{\partial L}{\partial b}= \sum_{i=1}^{l}(\beta_i-\alpha_i) = 0. \label{kkt2}\\
  & \hspace{-134mm} \frac{\partial L}{\partial \xi_i}=  \frac{C}{l}\tau - \alpha_i -\gamma_i = 0,  ~~ i=1 ,2,...,l.\label{kkt3} \\
& \hspace{-125mm}\frac{\partial L}{\partial \xi_i^*}=  \frac{C}{l}(1-\tau) - \beta_i -\lambda_i = 0, ~~ i=1 ,2,...,l.\label{kkt4}\\
& \hspace{-105mm} \frac{\partial L}{\partial \epsilon}= C\nu\tau(1-\tau) - (1-\tau) \sum_{i=1}^{l}\alpha_i - \tau \sum_{i=1}^{l}\beta_i -\eta = 0  \label{kkt epsilon}\\
& \hspace{-105mm} \alpha_i(y_i- (w^T\phi(x_i)+b)- (1-\tau)\epsilon -  \xi_i) =0, ~~ i=1 ,2,...,l. \label{kkt5}\\
 \hspace{-8mm}\beta_i((w^T\phi(x_i)+b)-y_i - \tau\epsilon - \xi_i^{*}) = 0, ~~ i=1 ,2,...,l.~~~~~~~\label{kkt6}\\
& \hspace{-140mm}  \gamma_i\xi_i = 0,~~ \lambda _i\xi_i^{*} = 0~~ i=1 ,2,...,l.\label{kkt7}\\
& \hspace{-180mm}  \eta \epsilon = 0. \label{kktepsilon1} \\
 & \hspace{-115mm} y_i- (w^T\phi(x_i)+b) \leq (1-\tau)\epsilon +  \xi_i, i=1 ,2,...,l.\label{kkt8} \\
& \hspace{-118mm} (w^T\phi(x_i)+b)-y_i \leq   \tau\epsilon + \xi_i^{*}, ~~ i=1 ,2,...,l., \label{kkt9}\\
 &\hspace{ -125mm} \epsilon \geq 0, ~\xi_i \geq 0,~~\xi_i^{*} \geq 0, ~~ i=1 ,2,...,l . \label{kkt10}
         \end{eqnarray}   
    Making the use the above KKT conditions, the Wolfe dual problem of the primal problem (\ref{esvqr_primal}) can be obtained as follows
     \begin{eqnarray}
    \min_{\alpha,\beta} \frac{1}{2}\sum_{i=1}^{l}\sum_{j=1}^{l}(\alpha_i- \beta_j)K(x_i,x_j)(\alpha_j-\beta_i) - \sum_{i=1}^{l}(\alpha_i-\beta_i)y_i  \nonumber\\
    & \hspace{-170mm}\mbox{subject to,} \nonumber \\
    &\hspace{-150mm} \sum_{i=1}^{l}(\alpha_i-\beta_i)= 0, \nonumber \\
     &\hspace{-120mm}   (1-\tau) \sum_{i=1}^{l}\alpha_i + \tau \sum_{i=1}^{l}\beta_i \leq  C\nu\tau(1-\tau), \nonumber \\
    & \hspace{-140mm}  0 \leq \alpha_i \leq \frac{C}{l}\tau,~~ i=~1,2,...l,       \nonumber \\
    & \hspace{-132mm} 0 \leq \beta_i \leq \frac{C}{l}(1-\tau), ~~i=~1,2,...l.  
    \label{nusvqrdual}   
    \end{eqnarray}
   The KKT conditions (\ref{kkt1})- (\ref{kkt10}) will help us to discover the various characteristics of the proposed $\epsilon$-SVQR model. At first, we shall state following preposition.
   \newline

  \textbf{Preposition 1.}  $~ \alpha_i\beta_i$=0 and $\xi_i \xi ^{*}_i$=0 holds $\forall$ $\textit{i=1,2,...l} $.\\ 
  
  Proof:- If possible, let us suppose there exists an index $i$ such that $\alpha_i\beta_i \neq 0$ holds. It implies that $\alpha_i\neq 0$ and $\beta_i \neq 0$. Therefore, from the KKT condition (\ref{kkt5}) and (\ref{kkt6}) we can obtain
  \begin{eqnarray}
  (y_i- (w^T\phi(x_i)+b)- (1-\tau)\epsilon -  \xi_i) =0  \label{11}\\
  \mbox{and}~~~~~~~~~~~~~~~~~~~~~~ ((w^T\phi(x_i)+b)-y_i - \tau\epsilon - \xi_i^{*}) =0. \label{22} 
  \end{eqnarray}
  Adding equation (\ref{11}) and (\ref{22}) gives
  $\xi_i^{*}+ \xi_i = -\epsilon$ which is  possible only when either $\xi_i < 0$ or $\xi_i^{*} < 0$. But, the KKT condition (\ref{kkt10}) requires $\xi_i \geq 0,~~\xi_i^{*} \geq 0, ~~for~ i=1 ,2,...,l.$ which contradicts our assumption. This proves $~ \alpha_i\beta_i$=0 $\forall$ $\textit{i=1,2,...l} $.
  
      On the similar line, let us suppose that there exists an index $i$ for which $\xi_i^{*} \xi_i\neq 0$. It means that $\xi_i \neq 0$ and $\xi_i^{*} \neq 0$ for which we can obtain $\gamma_i  = 0 $ and $ \lambda_i = 0$  from KKT condition (\ref{kkt7}).  For  $\gamma_i  = 0 $ and $ \lambda_i = 0$, 
      we will obtain $\alpha_i = \frac{c}{l}\tau$ and $\beta_i = \frac{c}{l}(1-\tau)$ from the KKT conditions (\ref{kkt3}) and (\ref{kkt4}) respectively, which is not possible as we have already proven that  $~ \alpha_i\beta_i$=0 $\forall$ $\textit{i=1,2,...l} $. This proves $~ \xi_i\xi_i^{*}$=0 $\forall$ $\textit{i=1,2,...l} $.

  \textbf{Preposition 2.} For all those data points $ (x_i,y_i)$, which lie inside or boundary of the  asymmetric $\epsilon$- insensitive tube, the corresponding $\xi_i$ and $\xi_i^{*}$ will take zero value. 
  
  Proof:- The data point $(x_i,y_i)$ lying inside or boundary of the asymmetric $\epsilon$- insensitive tube must satisfy
  \begin{eqnarray}
 y_i- (w^T\phi(x_i)+b)-  (1-\tau)\epsilon ~ \leq  0 \\
   \mbox{and} ~~(w^T\phi(x_i)+b)-y_i - \tau\epsilon~ \leq  0      
  \end{eqnarray}
   
   If possible, let us suppose that $\xi_i  \neq 0 $ which means that $\xi_i > 0 $
  (as the KKT condition (\ref{kkt10})
 requires $\xi \geq 0$). Since  $\xi_i > 0 $, we can obtain $\gamma_i =0 $ and further $\alpha_i  > 0 $ by using the KKT conditions (\ref{kkt7}) and (\ref{kkt5}) respectively. For $\alpha_i  > 0 $, the KKT condition (\ref{kkt5}) implies that
 \begin{equation}
 y_i- (w^T\phi(x_i)+b)- (1-\tau)\epsilon =  \xi_i,
 \end{equation}   
 which is not possible as $\xi_i > 0$.
 
                  On the similar line, we can show that $\xi_i^{*}$ also cannot take non-zero values.

   It is also easy to prove that data point, which lie outside of the asymmetric $\epsilon$- insensitive tube, the corresponding  $\xi_i$ or $\xi_i^{*}$ will take positive value. \\
   
  \textbf{Preposition 3.} For the data point $ (x_i,y_i)$, which lie insides of the asymmetric $\epsilon$- insensitive tube, the corresponding $\alpha_i$ and  $\beta_i$ will take zero value. 
  
  Proof:-  The data point $(x_i,y_i)$ lying inside of the $\epsilon$-tube ,the $\xi_i$ and $\xi_i^{*}$ = 0 which means
              \begin{eqnarray}
              y_i- (w^T\phi(x_i)+b)-  (1-\tau)\epsilon - \xi_i ~ \leq  0 ,\\
              \mbox{and} ~~(w^T\phi(x_i)+b)-y_i - \tau\epsilon -\xi^{*}_i~ \leq  0.      
              \end{eqnarray}
     For which the use of the KKT condition (\ref{kkt5}) and (\ref{kkt6}) will let us obtain $\alpha_i$ and  $\beta_i$=0.            
   
    \textbf{Preposition 4.}  For the data point $ (x_i,y_i)$, lying above of the $\epsilon$-tube, $\alpha_i = \frac{C}{l}\tau$ and $\beta_i=0$. For the data point $(x_i,y_i)$, lying below of the $\epsilon$-tube,  $\alpha_i=0$ and $\beta_i = \frac{C}{l}(1-\tau) $ .
    
     Proof:- The data point $ (x_i,y_i)$, lying above of the $\epsilon$-tube will hold
          \begin{eqnarray}
          y_i- (w^T\phi(x_i)+b)-  (1-\tau)\epsilon > 0,
          \end{eqnarray}
  for which the corresponding $\xi_i$ will take positive value for satisfying the KKT condition (\ref{kkt8}). For $\xi_i  >0$, we can get $\gamma_i=0$ from (\ref{kkt7}) and further can obtain $\alpha_i = \frac{C}{l}\tau $ from the KKT condition(\ref{kkt3}). Further $\beta_i$ will take zero value as  $\alpha_i\beta_i=0$.
  
                 On the similar line, we can prove that the data point $(x_i,y_i)$, lying below of the $\epsilon$-tube,  $\alpha_i=0$ and $\beta_i =  \frac{C}{l}(1-\tau)$ . 
                 
                   \textbf{Preposition 5.} For $ 0 <\alpha_i < \frac{C}{l}\tau $ ( $ 0 <\beta_i < \frac{C}{l}(1-\tau) $ ), the corresponding data point $(x_i,y_i)$ will be lying on the upper (lower) boundary of the asymmetric $\epsilon$- insensitive tube.  

    Proof:-  For  $ 0 <\alpha_i < \frac{C}{l}\tau $,  the $\gamma_i >0$ from KKT Condition (\ref{kkt3}) which implies $\xi_i = 0 $. Further, for $\alpha_i > 0 $ , we can obtain
     \begin{eqnarray}
    y_i- (w^T\phi(x_i)+b)  =  (1-\tau)\epsilon,
    \end{eqnarray}
           which means that data point $(x_i,y_i)$ will be lying on the upper boundary of the asymmetric $\epsilon$- insensitive tube. On the similar line, we can obtain that for $ 0 <\beta_i < \frac{C}{l}(1-\tau) $, the corresponding data point $(x_i,y_i)$ will be lying on the below boundary of the asymmetric $\epsilon$- insensitive tube.  
           
            Now, we can argue that the data point  $(x_i,y_i)$, which are lying outside of the $\epsilon$ -tube 
    
    \textbf{Remark 1.} For $\epsilon >0$ , the $\eta$ will take zero value and the inequalities constraint $(1-\tau) \sum_{i=1}^{l}\alpha_i + \tau \sum_{i=1}^{l}\beta_i \leq  C\nu\tau(1-\tau)$ of the dual problem (\ref{nusvqrdual}) will get converted to the equaltiy constraint 
    \begin{equation}
    (1-\tau) \sum_{i=1}^{l}\alpha_i + \tau \sum_{i=1}^{l}\beta_i =  C\nu\tau(1-\tau). \label{epsilon11}
    \end{equation}
    Now, we shall term  the data points which are lying outside of the asymmetric $\epsilon$-tube with `Errors'. The data points which are lying outside of the asymmetric $\epsilon$-tube as well as boundary of the tube is termed with the `support vectors'. These data points only contributes for the construction of the final regressor.
    \\
    
    \textbf{Preposition 6.} Suppose the $\nu$-SVQR is applied to some dataset and resulting $\epsilon$ is non zero then the follwing statements hold.
    \begin{enumerate}
    	\item   [(a)] $\nu$ is upper bound on the fraction of of Errors.
    	 \item [(b)] $\nu$ is lower bound on the fraction of of Support vectors
    	 .
    \end{enumerate}
   
     Proof:- Let us suppose that there are $m_1$ and $m_2$ data points which are lying above and below of the asymmetric $\epsilon$-tube respectively.
     For the data point, lying above of the asymmetric $\epsilon$-tube, only $\alpha_i$ will take the value $\frac{C}{l}\tau $. For the data point, lying below of the asymmetric $\epsilon$-tube, only $\beta_i$ will take the value $\frac{C}{l}(1-\tau)$.  The data point $(x_i,y_i)$ lying on the upper (lower) boundary of the asymmetric $\epsilon$- insensitive tube will be taking $ 0 <\alpha_i < \frac{C}{l}\tau $ ( $ 0 <\beta_i < \frac{C}{l}(1-\tau) $ ) values.

      For $\epsilon >0 $ ,  we  can obtain from \ref{epsilon11},
       \begin{eqnarray*}
       m_1\frac{C}{l}(1-\tau)\tau + m_2 \frac{C}{l}(1-\tau)\tau ~~ \leq ~~  C\nu\tau(1-\tau)
       \end{eqnarray*}
     which implies that  $\frac{1}{l}(m_1+ m_2) ~ \leq \nu $.

      Furthermore, there should exist at least $m_1$ and $m_2$ data points lying above and below of the asymmetric $\epsilon$-tube which would satisfy the equality \ref{epsilon11}.
     For these data points we have,
       \begin{eqnarray*}
      	m_1\frac{C}{l}(1-\tau)\tau + m_2 \frac{C}{l}(1-\tau)\tau ~~ = ~~  C\nu\tau(1-\tau)
      \end{eqnarray*} which implies that  $\frac{1}{l}(m_1+ m_2) ~ = \nu $.  It further means that there should at least $\nu$ fraction of the support vectors.
  
   \textbf{Remark 2.} Asymptotically, the $\nu$ equals the fraction of support vectors and errors.  The probability of the data point lying on the boundary of the asymmetric $\epsilon$-tube becomes zero asymptotically. This statement can be proved under certain condition similar to the proof of the Prepostion  1 (iii)  given in ( Scholkopf, \cite{newsvr} ). But however, in this paper we shall empirically verify that the $\nu$ equals the fraction of support vectors and errors asymptotically.
   
   \textbf{Remark 3.}  Asymptotically, the data points appear above and below of the asymmetric $\epsilon$-tube in the ratio of $1-\tau$ and $\tau$ respectively in the $\nu$-SVQR model. It means that for the large value of $l$, there would be $l\nu(1-\tau)$ and $l\nu\tau$  data points lying above and below of the asymmetric $\epsilon$-tube respectively. It is because of the facts that $\alpha_i$ and $\beta_i$ also have to satisfy the KKT condition (\ref{kkt2}).
   
   \textbf{Remark 4.} If proposed $\nu$-SVQR  obtains the solution $(\bar{w}$,$\bar{b}$ $\bar{\epsilon})$ with parameter value $C'$, then $\epsilon$-SVQR model with parameters $\epsilon=\bar{\epsilon}$ and
 $C=C'*N$ will obtain the same solution $\bar{w}$,$\bar{b}$.

      \underline{Obtaining the value of $\epsilon$ and $b$} :- At first we  can obtain the value of the $\epsilon$ which is the effective width of the asymmetric tube. For this, we find out the data point which are lying on the upper and lower boundary of the $\epsilon$-tube using the preposition-5. For  $ 0 <\alpha_i < \frac{C}{l}\tau$, we can obtain the upper width of the asymmetric tube  using $ y_i- (w^T\phi(x_i)+b))$.  For $ 0 <\beta_j < \frac{C}{l}(1-\tau)$,  we can obtain the lower width of the asymmetric tube  using   $(w^T\phi(x_j)+b)-y_j$.  But, the computation of final width of the asymmetric $\epsilon$-tube does not require the value of $b$  and can be obtained by
      \begin{eqnarray}
       \epsilon = (y_i- (\sum_{k=1}^{l} (\alpha_k-\beta_k)K(x_k,x_i))) + (\sum_{k=1}^{l} (\alpha_k-\beta_k)K(x_k,x_j))-y_j), \nonumber  \\
       & \hspace{-140mm} \mbox{where}~~ 0 <\alpha_i < \frac{C}{l}\tau ~~\mbox{and}~~ 0 <\beta_j < \frac{C}{l}(1-\tau). 
      \end{eqnarray}
                          After obtaining the value of $\epsilon$, we can obtain the value of $b$. For $ 0 <\alpha_i < \frac{C}{l}\tau$ , we can obtain
                          \begin{eqnarray}
                          b= y_i-\sum_{k=1}^{l} (\alpha_k-\beta_k)K(x_k,x_i)-(1-\tau)\epsilon \label{b1}
                          \end{eqnarray}
                          For $ 0 <\beta_j < \frac{C}{l}(1-\tau)$ , we can also obtain
                          \begin{eqnarray}
                          b= y_j -\sum_{k=1}^{l} (\alpha_k-\beta_k)K(x_k,x_j)+ \tau\epsilon \label{b2}
                          \end{eqnarray} 
            In practice,  we compute  values of $b$ form equation (\ref{b1}) and (\ref{b2}) and use their average value as the final value of $b$. After computing the values of decision variables $\alpha$, $\beta$ and $b$ the quantile regression is estimated by
            \begin{equation}
            f_\tau(x) = \sum_{i=1}^{l}(\alpha_i-\beta_i)K(x,x_i) +b
            \label{r12}
            \end{equation}

            Further, like $\epsilon$-SVR model described in (Gunn, \cite{GUNNSVM}), if the kernel contains a bias term then, the $\nu$-SVQR dual problem (\ref{nusvqrdual}) can be solved without equality constraint and  the quantile regression function is simply estimated by
            \begin{equation}
            f_\tau(x) = \sum_{i=1}^{l}(\alpha_i-\beta_i)K(x,x_i) 
            \label{r122}
            \end{equation}

       \section{Experimental Section}
       In this section, we shall empirically verify the claims made in this paper. For this, we first describe our experimental setup. We have performed all experiments with MATLAB 17.0 environment (http://in.mathworks.com/) on Intel i7 processor with 8.0 GB of RAM. The QPPs of proposed $\nu$-SVQR and $\epsilon$-SVQR has been solved by the quadprog function with interior-point convex algorithm available in the MATLAB 16.0 environment.
       For all of the experiments, we have used the RBF kernel function $exp(\frac{-||x-y||^2}{q})$, where $q$ is the kernel parameter and quantile regression function is estimated by  (\ref{r12}). 
       The proposed $\nu$-SVQR model requires three parameters to be tunned namely RBF kernel parameter $q$,$C$ and $\nu$ where as the  $\epsilon$-SVQR model requires the tunning of  parameters $q$,$C$ and $\epsilon$. All these parameters have been tunned using exhaustive search method (Hsu and Lin, \cite{Exhaustivesearch}). The parameter $q$ and $C$ has been searched in the set $\{ 2^i: i=-15,-9,......9,15\} $. 
       
       \subsection{\textbf{Performance Criteria}}  \label{Perform_criteria}                                   
       For the evaluation of the efficacy of  SVQR  models, we have used some evaluation criteria which is also mentioned in (Xu Q et al., \cite{Weighted_QSVR}).
       Given the training set $T= \{ (x_i,y_i): x_i \in \mathbb{R}^n, y_i \in \mathbb{R},~ i=1,2...,l~ \}$  and true  $\tau$-th conditional quantile function $Q_{\tau}(y/x)$, we list the evaluation criteria as follows.  
       \begin{enumerate}
       	\item[(i)] $ RMSE$: It is Root Mean Square of Error.\\ ~~It is given by $\sqrt{ \frac{1}{l}\sum_{i=1}^{l}( Q_{\tau}(y_i/x_i)- f_\tau(x_i))^2}$.
       	\item[(ii)] $ MAE$: It is  Mean of the Absolute Error. \\
       	~~~It is given by ${ \frac{1}{l}\sum_{i=1}^{l}|( Q_{\tau}(y_i/x_i)- f_\tau(x_i))|}$.
       	\item[(iii)]  Error $E_\tau$: It is the measure which is used when the true quantile function is unknown.
       	It is given by $E_\tau ~=~ |p_\tau -\tau|$, where  $p_{\tau} = P(y_i \leq f_{\tau}(x_i))$ is the coverage probablity. For the real world UCI datasets  experiments, we would be using this measure. We shall compute the coverage probability $p_{\tau}$ by obtaining the estimated $\tau$ value in 100 random trails.
       	\item[(iv)]  Sparsity(u) = $\frac{\#(u=0)}{\#(u)}$, where $\#(r)$ determines the number of the component of the vector $r$ .
       \end{enumerate}

       \subsection{Artifical Datasets}
       We shall show different properties of proposed $\nu$-SVQR model and its advantages over $\epsilon$-SVQR model  empirically. The best way to do this is to generate artificial datasets as actual true quantile can be easily computed for these datasets and unbiased comparisons can be made.  We have generated the training set $T$ where $x_i$ is drawn from the univariate uniform distribution with  $[-4,4]$.  The response variable $y_i$ is obtained from polluting a nonlinear function  of $x_i$ with different natures of noises in artificial datasets as follows.
       \begin{eqnarray}
       & \hspace{5mm} \mbox{AD1:}~~y_i = (1-x_i+2x_i^{2})e^{-0.5x_{i}^2} + \xi_i, \nonumber
       \mbox{~~~~~where $\xi_i$ is from N(0, $\sigma$).}  \nonumber\\
       & \hspace{5mm} \mbox{AD2:}~~y_i = (1-x_i+2x_i^{2})e^{-0.5x_{i}^2} + \xi_i, \nonumber
       \mbox{~~~~~where $\xi_i$ is from U(a, b).}  \nonumber
       \end{eqnarray}
       The true quantile function $Q_\tau(y_i/x_i)$ in these artificial datasets can be obtained as
       \begin{eqnarray}
       y_i = (1-x_i+2x_i^{2})e^{-0.5x_{i}^2} + F_{\tau}^{-1}(\xi_i), \nonumber
       \end{eqnarray}  
       where $F_{\tau}^{-1}(\xi_i)$ is the  $\tau$th quantile of random error $\xi_i$. We have evaluated the SVQR models by generating 1000 testing points in each trails.
       
       \subsection*{ Experiment 1}
\begin{figure}
    	\centering
    	\begin{tabular}{cc}
    		\subfloat[] {\includegraphics[width = 2.5in,height=1.8in]{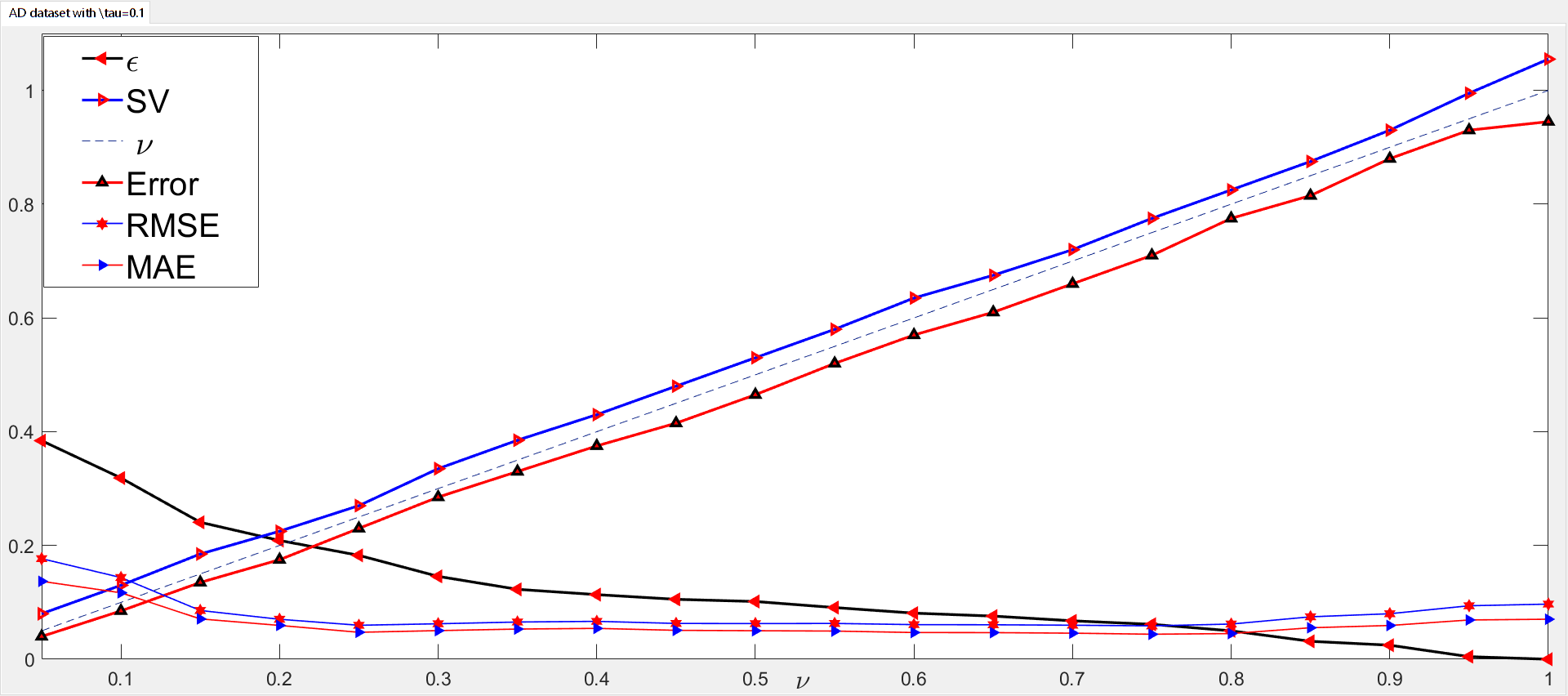}}
    		\subfloat[] {\includegraphics[width = 2.5in,height=1.8in]{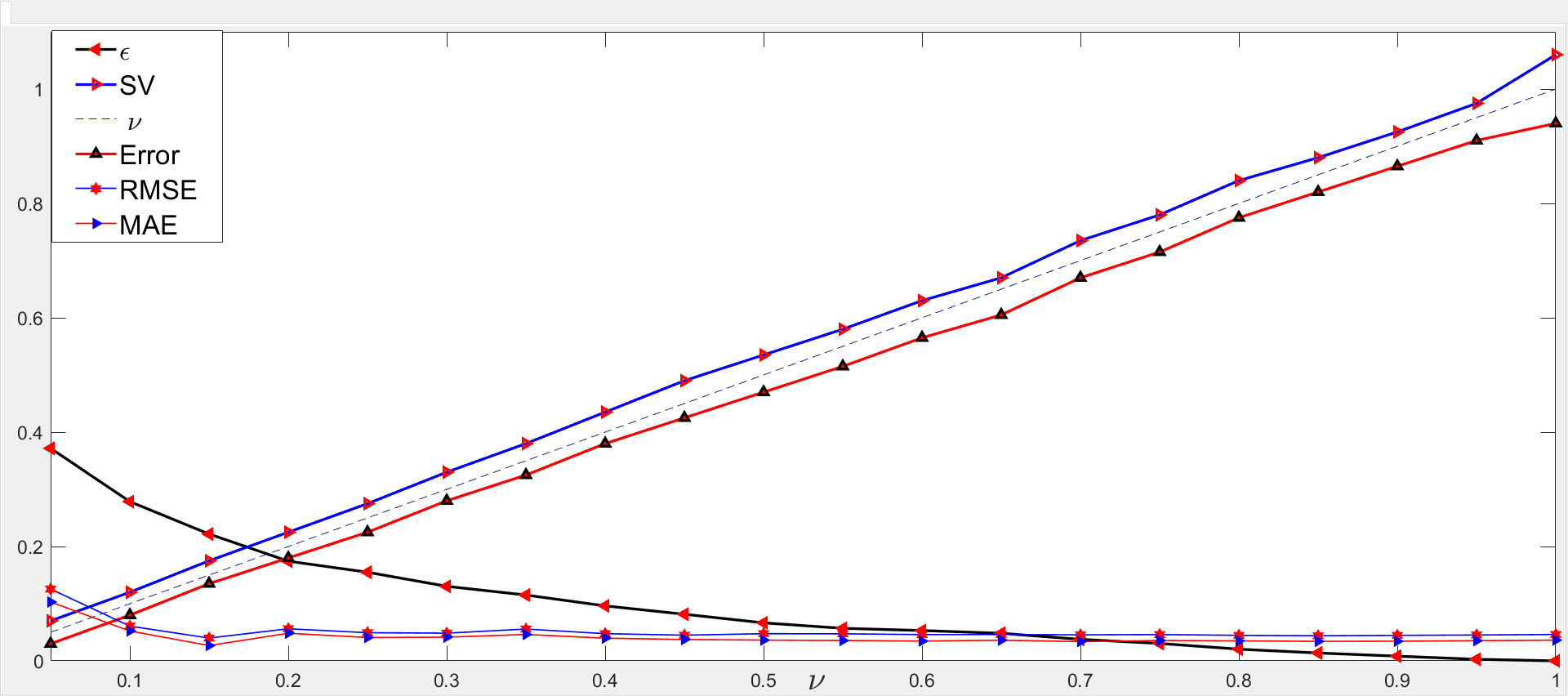}} \\
    		\subfloat[] {\includegraphics[width = 2.5in,height=1.8in]{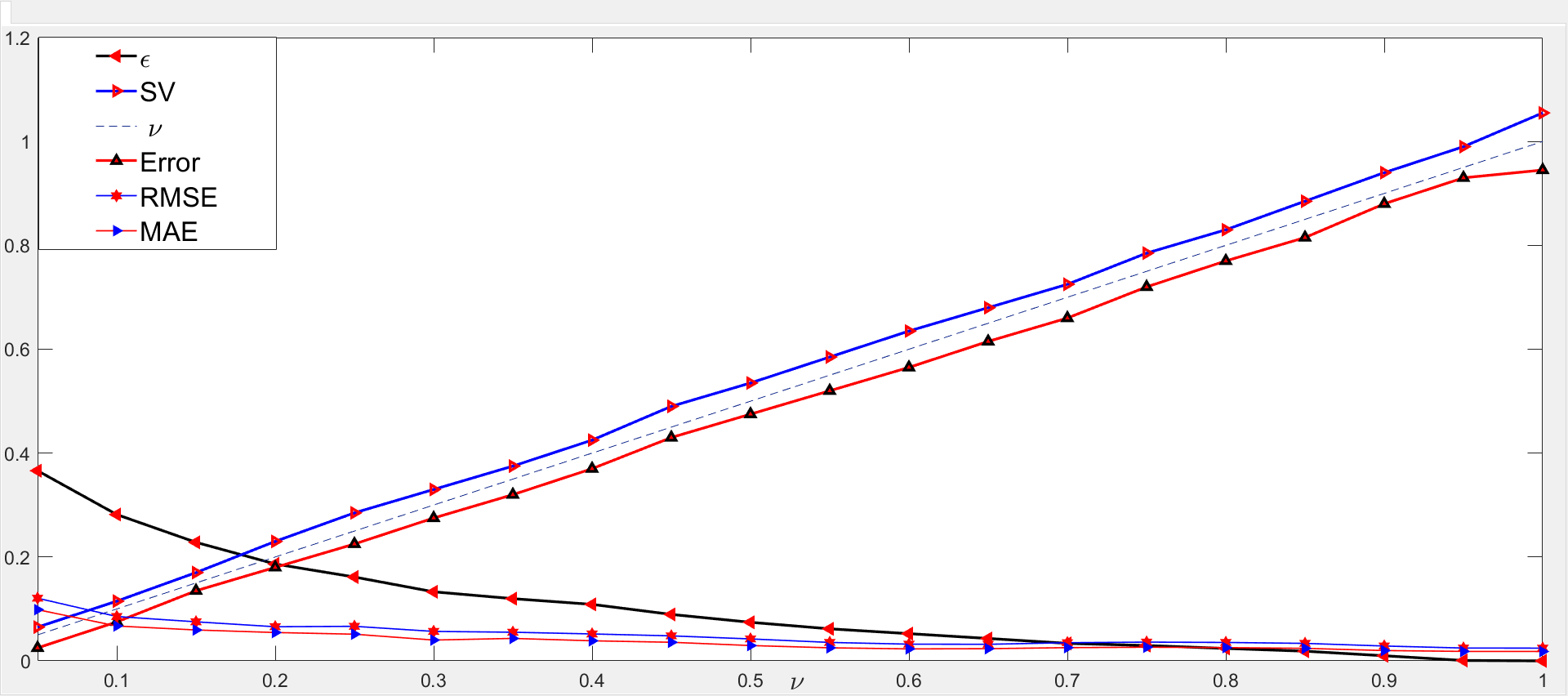}}
    		\subfloat[] {\includegraphics[width = 2.5in,height=1.8in]{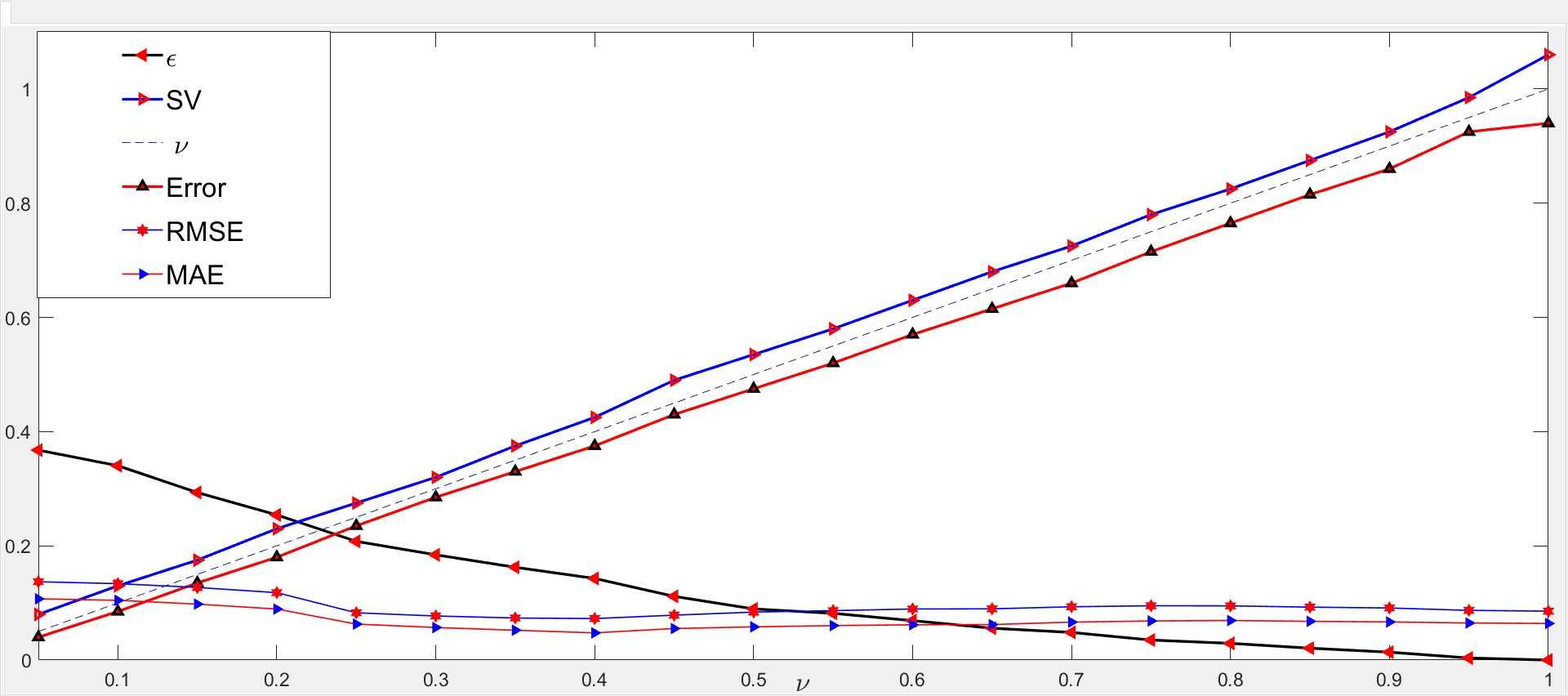}}\\
    	\end{tabular}
    	\caption{Performance of the proposed $\nu$-QSVR model on AD1 dataset  for (a) $\tau$=0.1 (b)  $\tau$=0.3 (c)  $\tau$=0.6  (d)  $\tau$=0.9.}
    	\label{plot1}
    	
    \end{figure}

        Our first experiment will empirically verify the Preposition-6 of this paper. For this, we have generated 200 training data points of AD1 artificial  datasets and obtain the  numerical results for 10 random simulations. Table \ref{table1} shows the performance of the  proposed $\nu$-SVQR model on several values of $\nu$ for  $\tau$= 0.2 ,0.5,0.7 and 0.8. Figure \ref{plot1} shows the proposed $\nu$-SVQR model on several values of $\nu$ for  $\tau$= 0.1 ,0.3,0.6 and 0.9. Following observation can be easily drawn form numerical results listed in the Table\ref{table1} and plots of Figure \ref{plot1}.
        \begin{enumerate}
        	\item[(a)]  Irrespective of $\tau$ values, as the value $\nu$ increases, the total width $\epsilon$ of asymmetric $\epsilon$-insensitive zone decreases. 
        	\item[(b)] Irrespective of $\tau$ values, $\nu$ is upper bound on the fraction of of Errors.
        	\item[(c)] Irrespective of $\tau$ values, $\nu$ is lower bound on the fraction of of support vectors.        
        	\item [(d)] RMSE and MAE obtained by proposed $\epsilon$-SVQR model also varies with the parameter $\nu$.
        \end{enumerate}
        
       \subsection*{Experiment 2}  
\begin{figure}
    	\centering
    	\begin{tabular}{cc}
    		\subfloat[] {\includegraphics[width = 4.2in,height=1.5in]{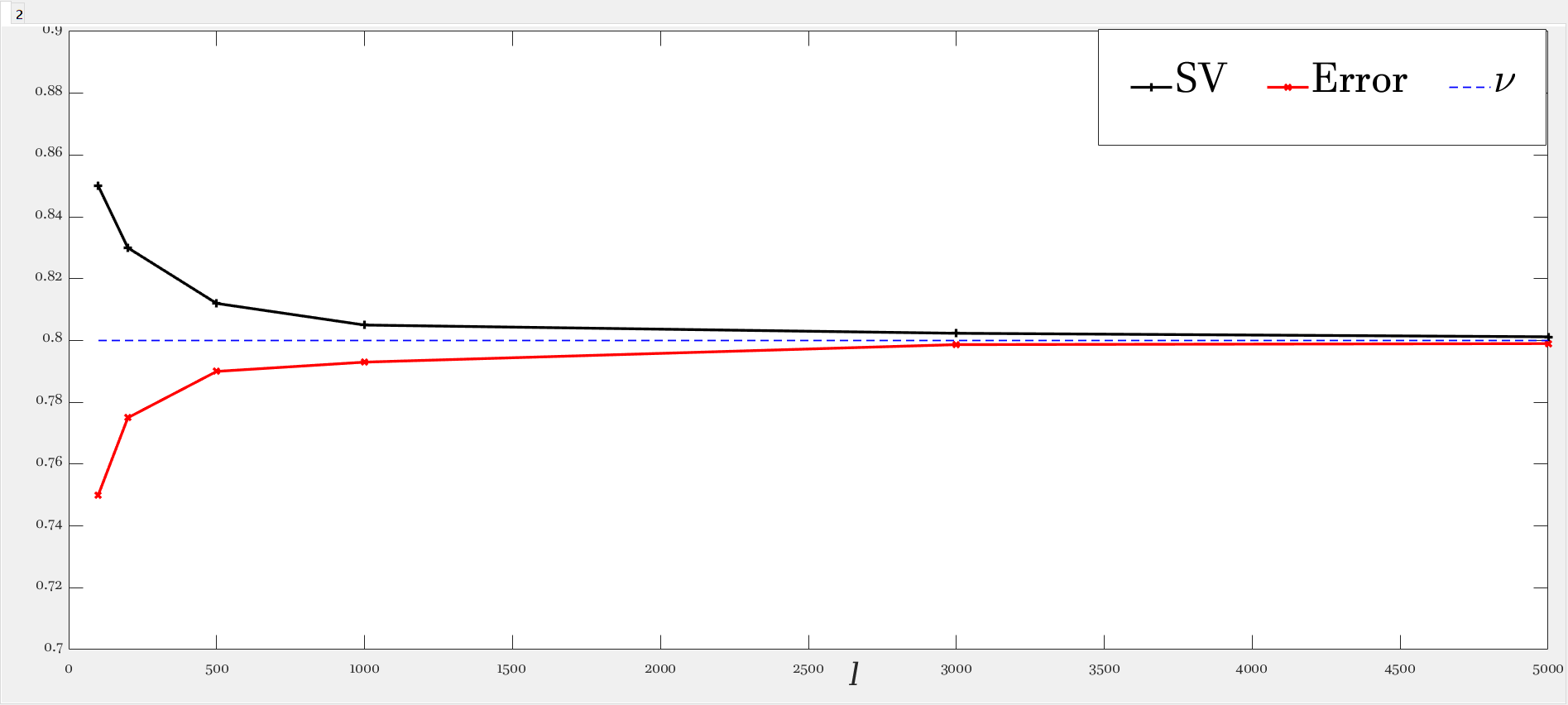}}\\
    		\subfloat[] {\includegraphics[width = 4.2in,height=1.5in]{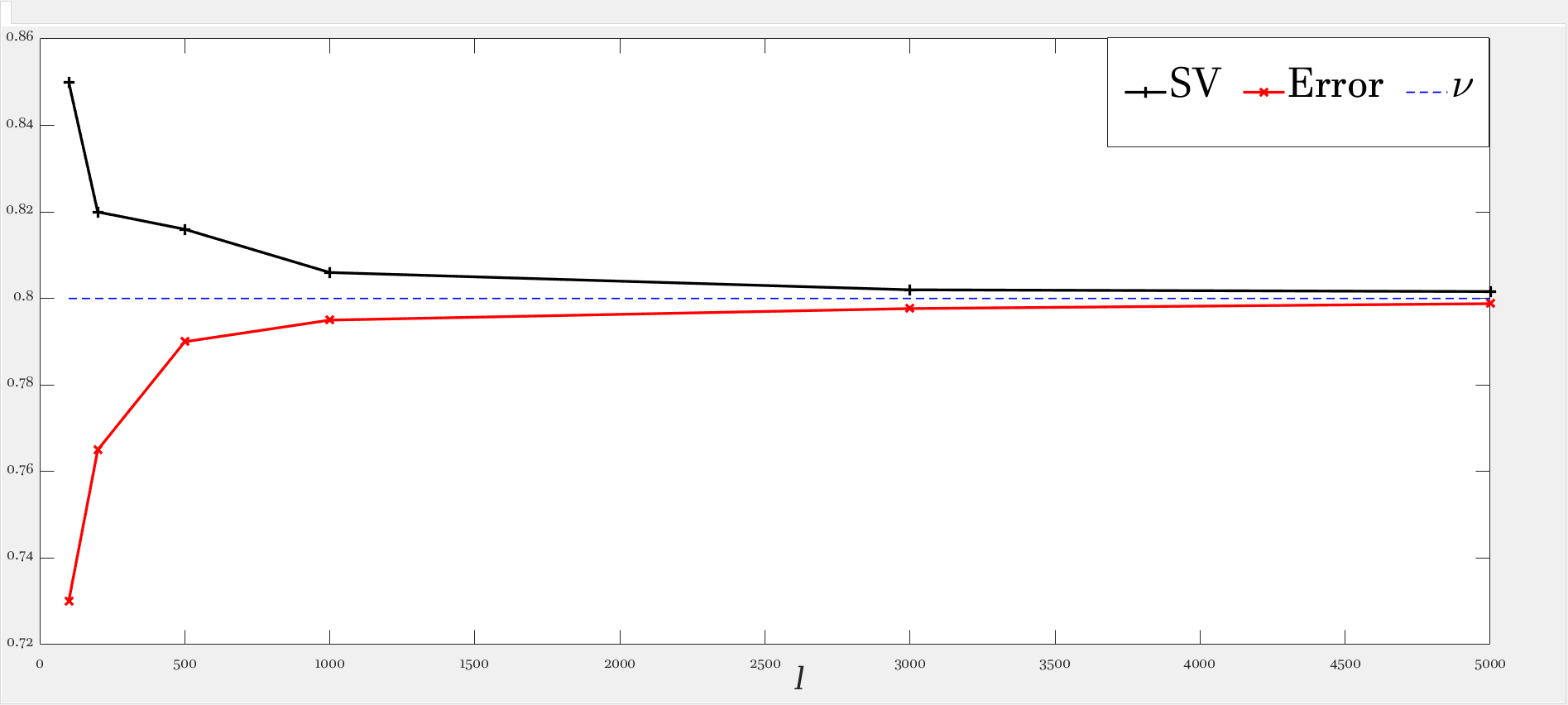}}\\
    		\subfloat[] {\includegraphics[width = 4.2in,height=1.6in]{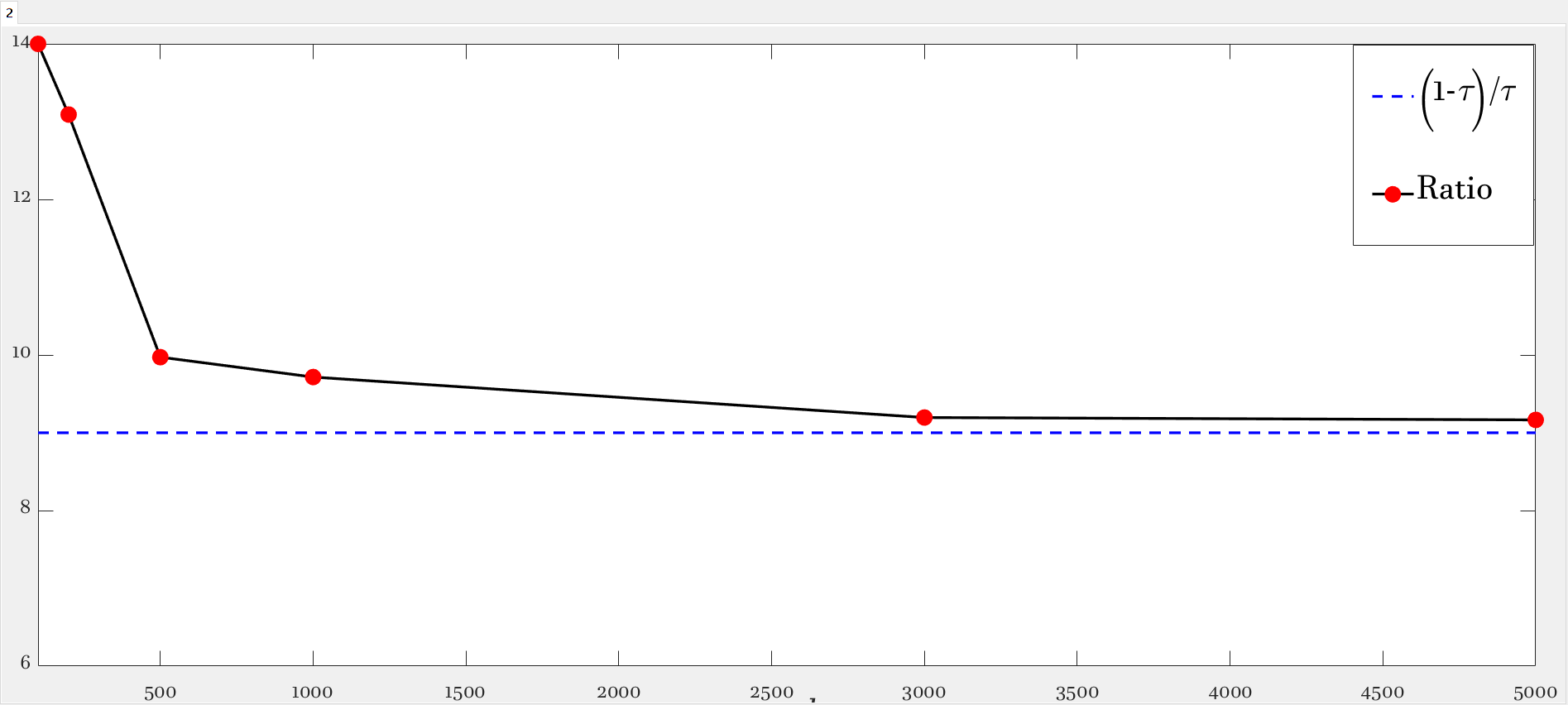}}\\
    		\subfloat[] {\includegraphics[width = 4.2in,height=1.6in]{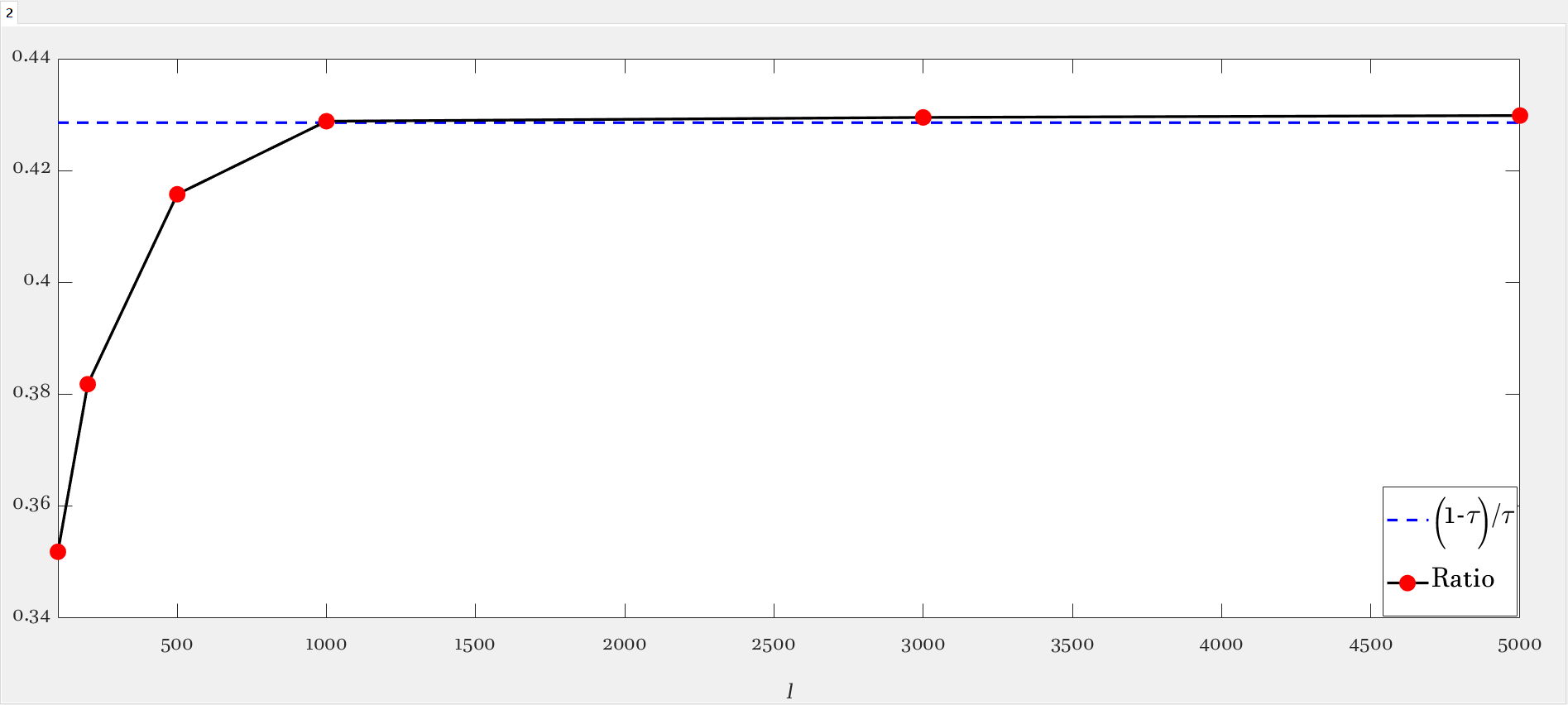}}
    	\end{tabular}
    \caption{ \\Asymptotic behavior of $\nu$-SVQR model for (a) $\tau=0.1$ (b) $\tau=0.3$ (c) $\tau=0.1$ and (d) $\tau=0.7$}
    \label{asymptotic}
\end{figure}

        The second experiment has been performed with the varying number of training points of AD1 dataset for observing the asymptotic behavior of proposed $\nu$-SVQR model.  For this experiment, we have fixed the $\nu =0.8$ in proposed $\nu$-SVQR model. Table \ref{table2} results the numerical results obtained by proposed $\nu$-SVQR model on AD1 dataset with different size of training set. In this Table, 'ratio' is the ratio of training data points lying above and below the asymmetric $\epsilon$-insensitive tube.
        Following facts can be easily observed from the numerical results listed in Table \ref{table2}.
        \begin{enumerate}
        	\item[(a)] Irrespective of $\tau$ values, the fraction of support vectors and errors converges to the $\nu$ value in the proposed $\nu$-SVQR model. It has also been well illustrated by the plot in Figure \ref{asymptotic}(a) and  \ref{asymptotic}(b). It is only because of fact that the probability of a training data point lying on boundaries of asymmetric $\epsilon$ -insensitive tube vanishes, as the number of training point increases.
        	\item[(b)] Irrespective of $\tau$ values, the ratio of training data point lying above and below of the asymmetric $\epsilon$-tube converges to $\frac{(1-\tau)}{\tau}$. It has also been well illustrated by the plot in Figure \ref{asymptotic}(c) and  \ref{asymptotic}(d). It means that the asymmetric $\epsilon$-insensitive zone used in proposed $\nu$-SVQR model is very suitable for handling quantile estimation problem.
        	\item [(c)] The resulting overall width of $\epsilon$-insensitive zone converges to a constant value in the proposed $\nu$-SVQR model.
        	\item [(d)] As the number of training points increases, there are more information available to the proposed $\nu$-SVQR model. It results in decrease in RMSE values obtained by proposed $\nu$-SVQR model.
        	
        \end{enumerate}
        
        \subsection*{Experiment 3} 
\begin{figure}
	\centering
	\begin{tabular}{cc}
		\subfloat[] {\includegraphics[width = 4.0in,height=1.5in]{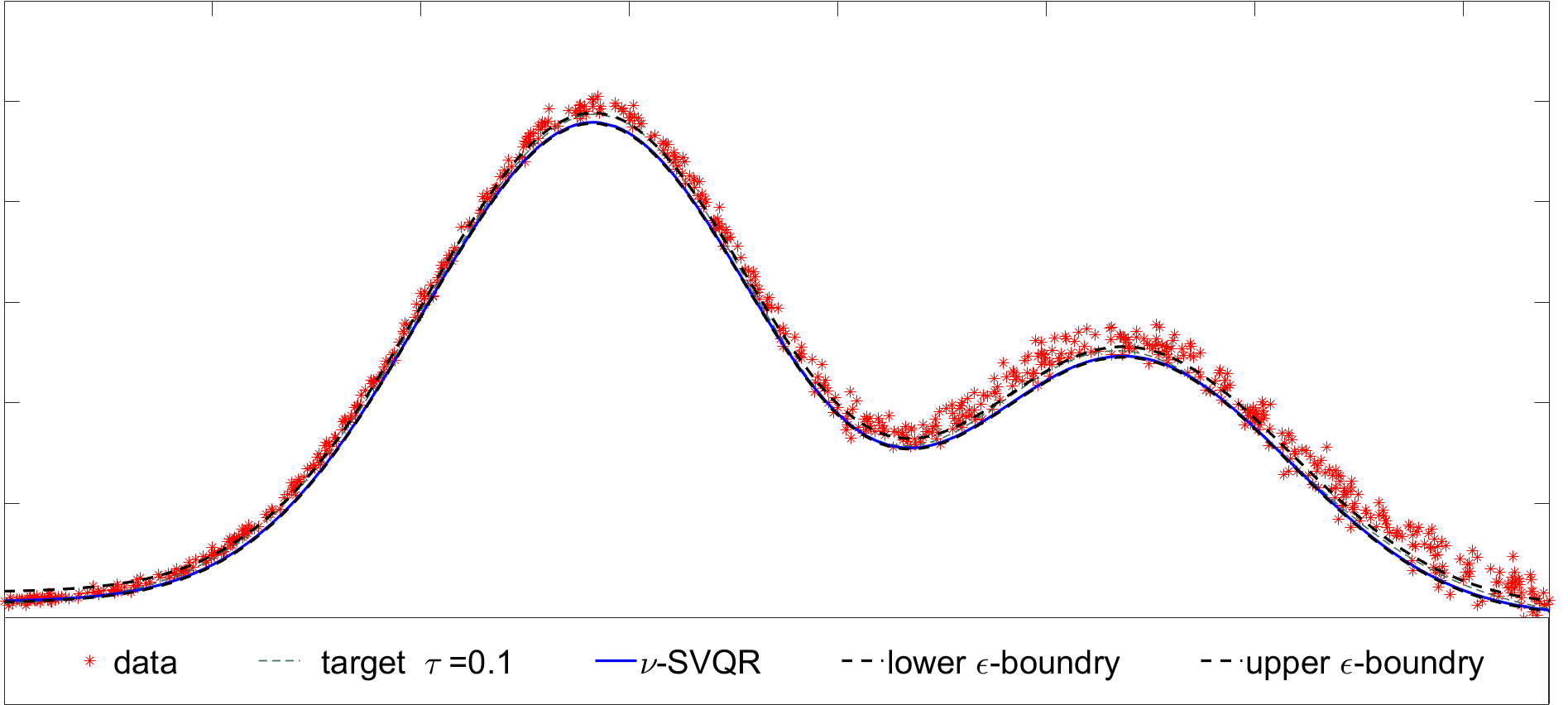}}\\
		\subfloat[] {\includegraphics[width = 4.0in,height=1.5in]{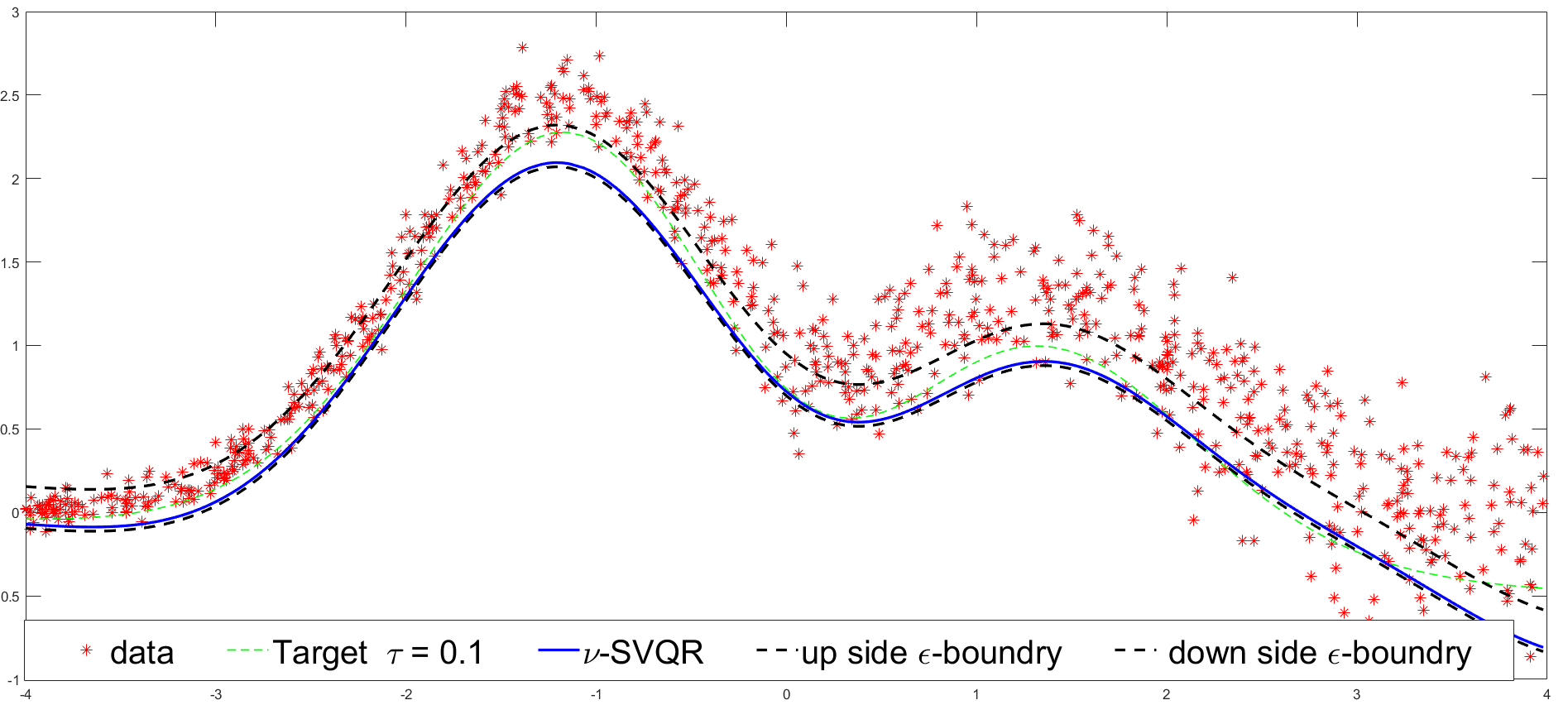}}\\
		\subfloat[] {\includegraphics[width = 4.0in,height=1.6in]{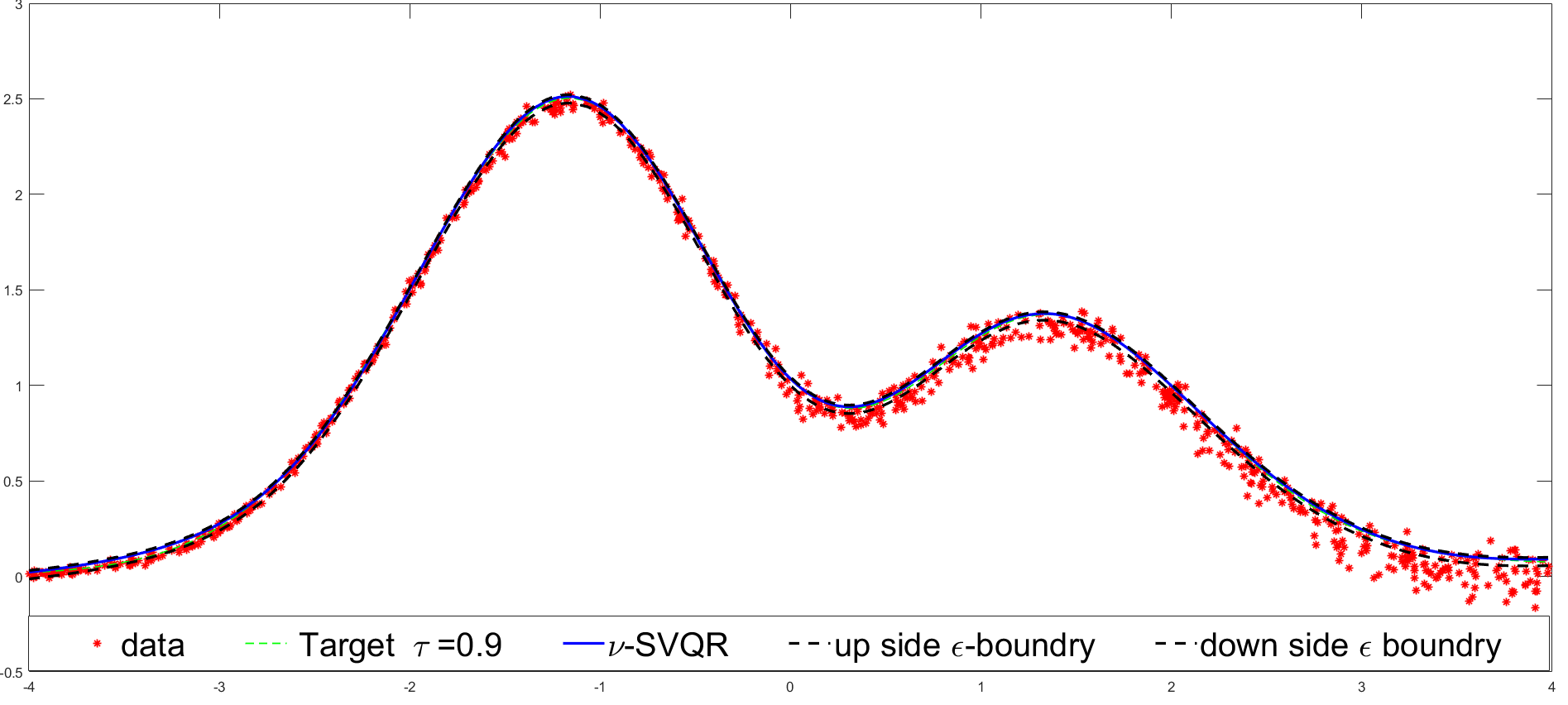}}\\
		\subfloat[] {\includegraphics[width = 4.0in,height=1.6in]{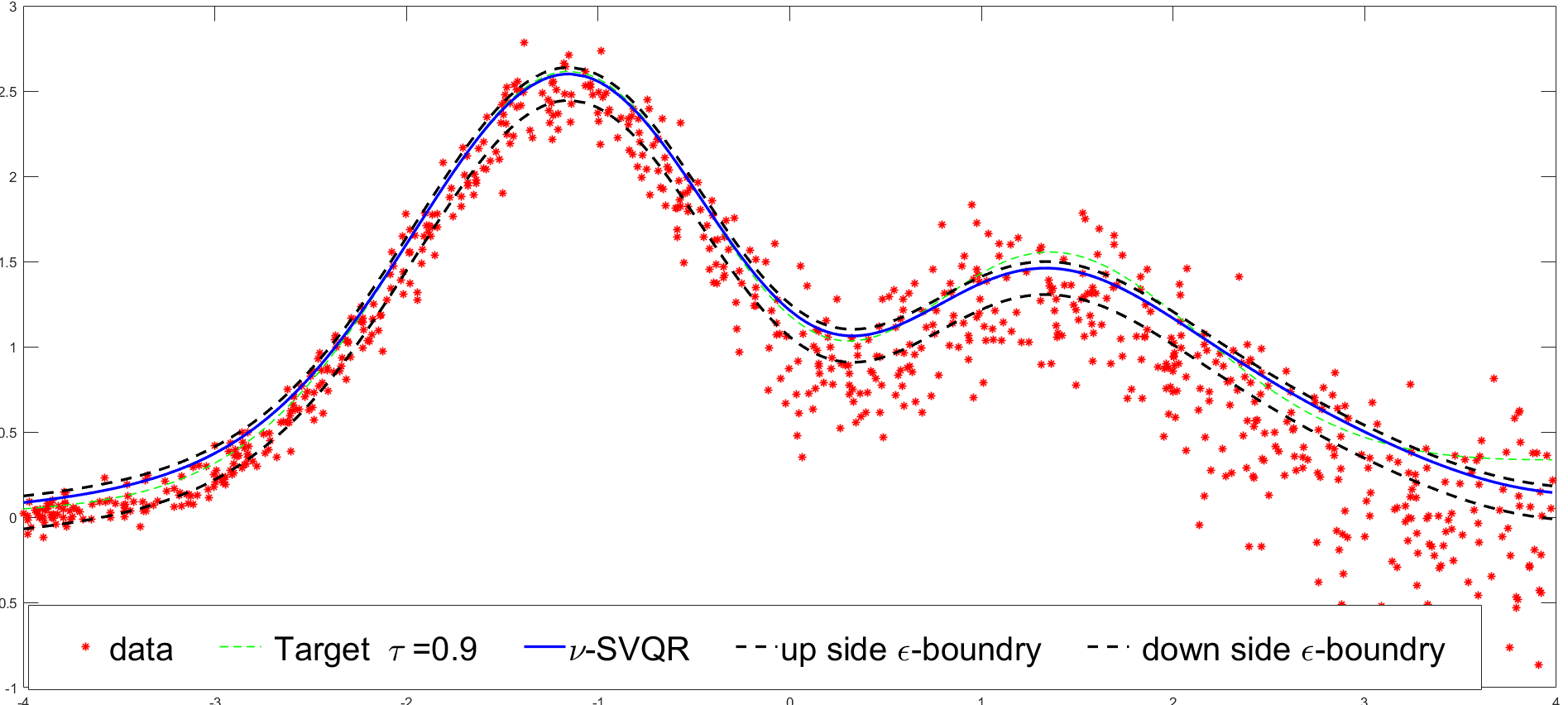}}
	\end{tabular}
	\caption{Automatic adjustment of the width of the $\epsilon$-insensitive zone in proposed $\nu$-SVQR model}
	\label{automaticadjust}
\end{figure}

This experiment has been performed to show the capability of the proposed $\nu$-SVQR model to automate the control over accuracy.  The proposed $\nu$-SVQR model has capability to automatically adjust the width of the asymmetric $\epsilon$-insensitive zone for efficient prediction.  For fix values of parameters with $\nu=0.3$, we have simulated the proposed $\nu$-SVQR model on AD1 dataset with noise variance $\sigma= 0.2$ and $\sigma=1$. Figure \ref{automaticadjust} shows the estimates obtained by proposed $\nu$-SVQR along with the $\epsilon$-insensitive zone for $\tau=$0.1 and 0.9 at fixed value of $\nu=0.3$. It can be observed that the proposed $\nu$-SVQR model can automatically adjust the width of the asymmetric $\epsilon$-insensitive zone according to the variance present in data for obtaining efficient estimates of quantiles.
        
\begin{figure}
	\centering
	\begin{tabular}{cc}
		\subfloat[] {\includegraphics[width = 2.5in,height=1.5in]{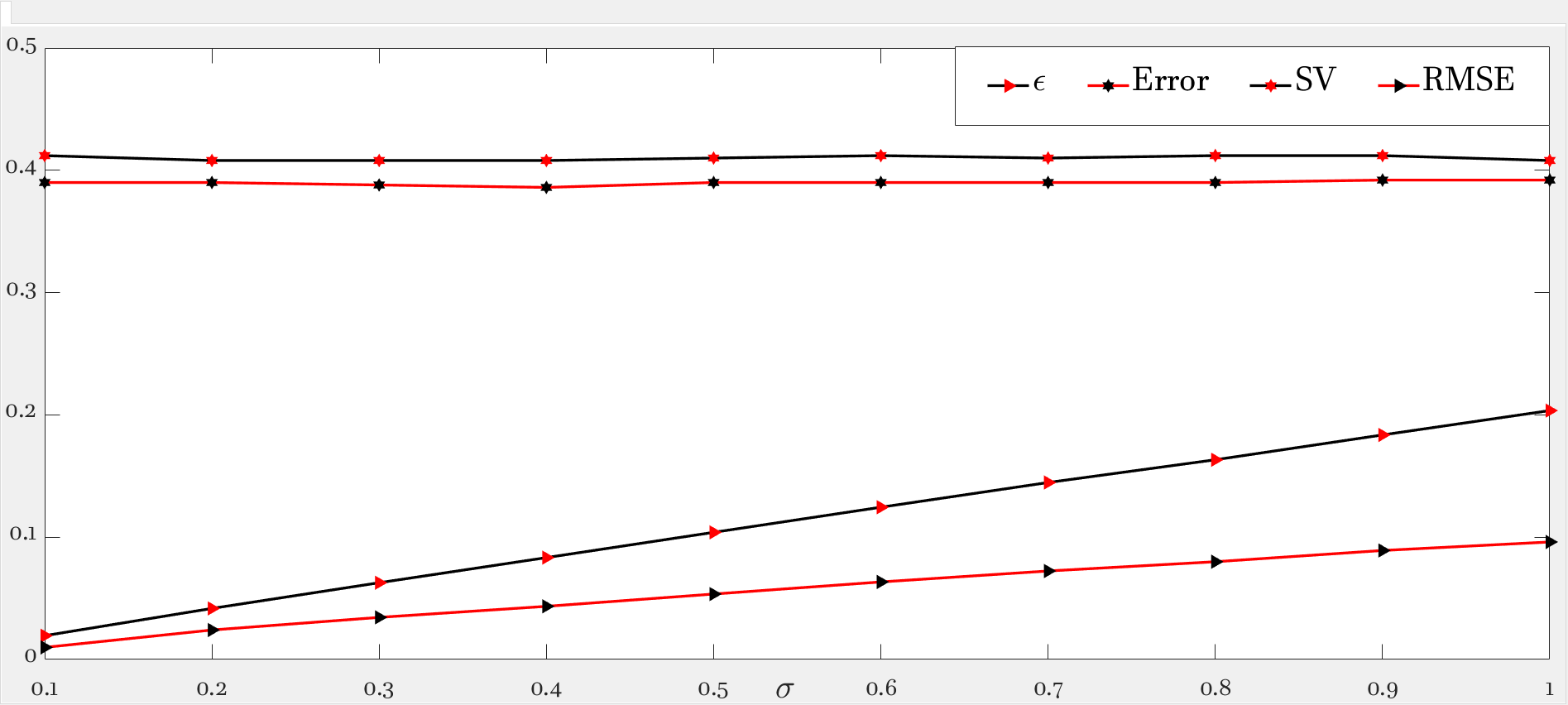}}
		\subfloat[] {\includegraphics[width = 2.5in,height=1.5in]{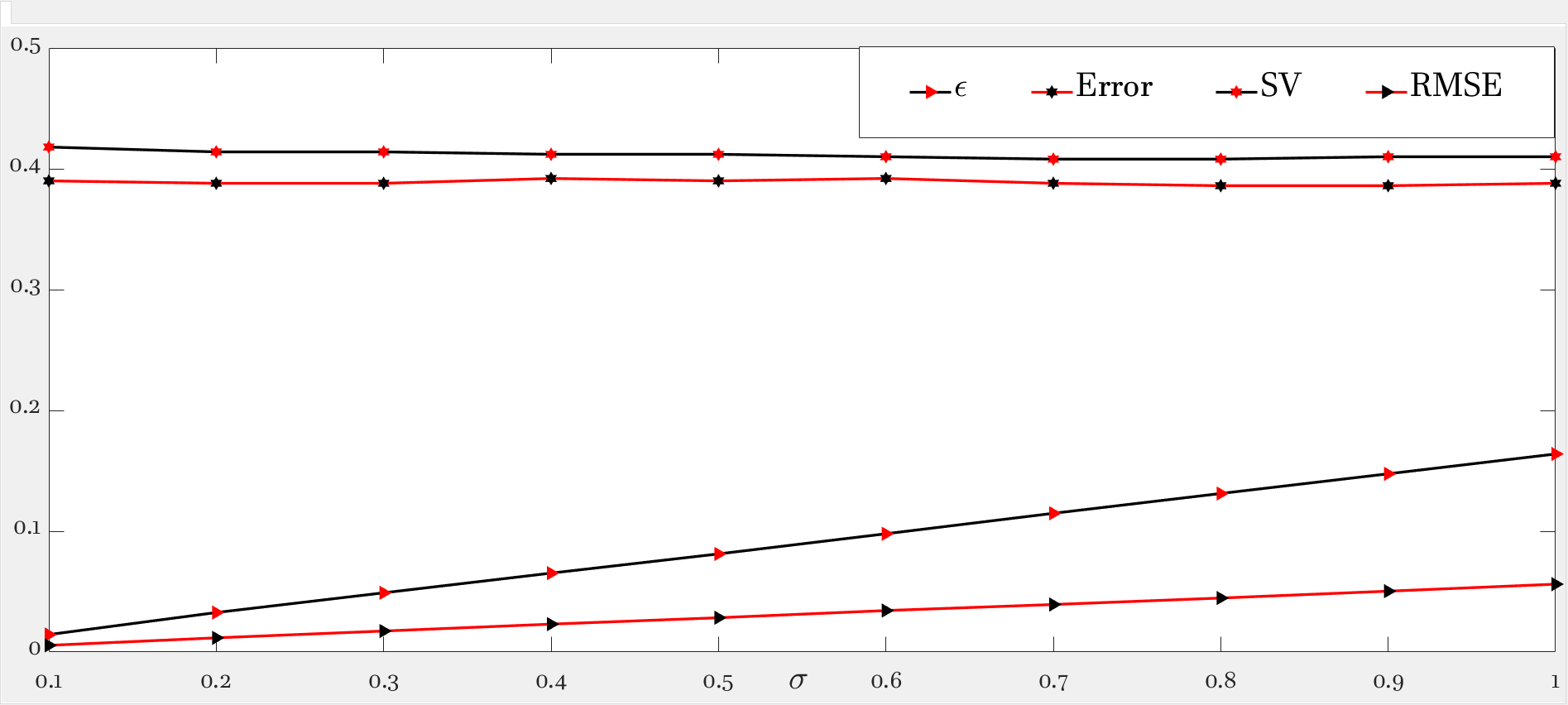}}\\
		\subfloat[] {\includegraphics[width = 2.5in,height=1.6in]{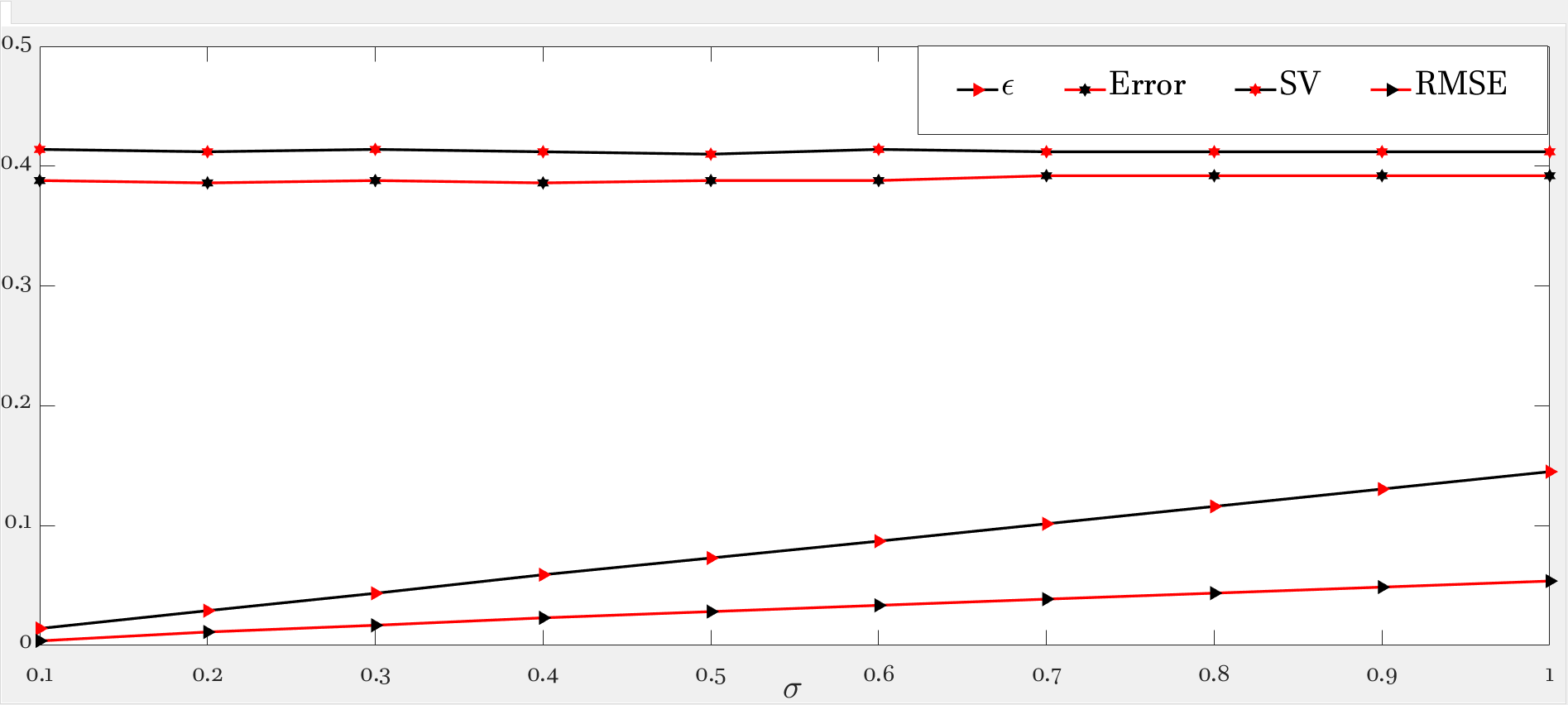}}
			\subfloat[] {\includegraphics[width = 2.5in,height=1.6in]{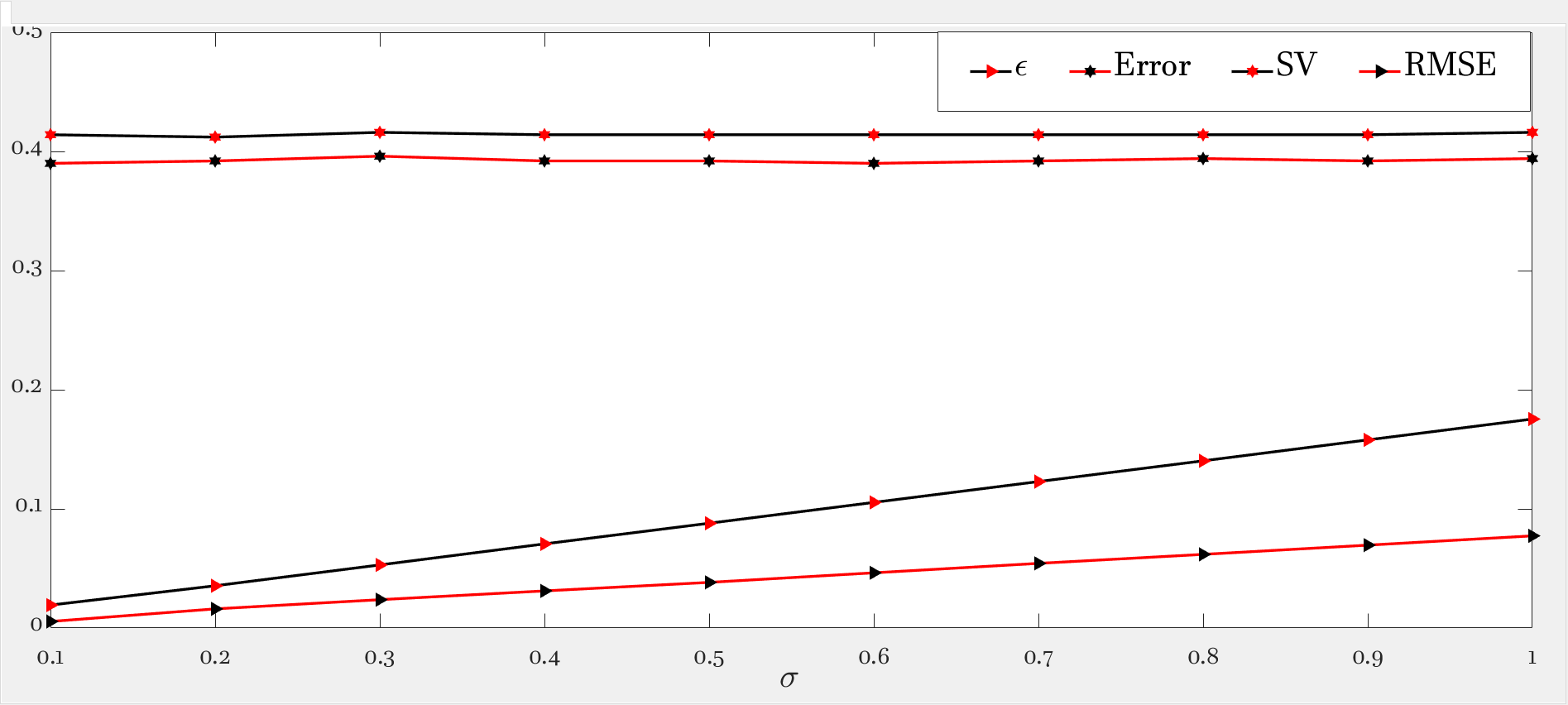}}
	\end{tabular}
	\caption{ The proposed $\nu$-SVQR model with different noise variance $\sigma$ for (a) $\tau=0.1$ (b) $\tau=0.3$ (c) $\tau=0.7$ and (d) $\tau=0.9$}
	\label{sigmaval}
\end{figure}

         Further, we have checked the performance of the proposed $\nu$-QSVR model on AD1 dataset with  different noise variance $\sigma$. For this experiment, we have fixed the number of training data points to 500. The $\nu$ parameter in proposed $\nu$-QSVR was fixed to 0.4. Other parameters were also fixed. Table \ref{table3} lists numerical results obtained by the proposed $\nu$-SVQR model on AD1 dataset with different noise variance $\sigma$ for several $\tau$ values. Figure \ref{sigmaval} illustrates the  numerical results listed in the Table \ref{table3} well for some $\tau$ values.  Following things can be easily observed.
         \begin{enumerate}
         	\item [(a)] Irrespective of values of $\tau$, as the noise variance $\sigma$ increases, the $\nu$-SVQR model accordingly increases the width of the asymmetric $\epsilon$-insensitive zone.
         	\item[(b)] Irrespective of values of $\sigma$, $\nu$ is the a upper bound on fraction of errors and lower bound on fraction of support vectors in proposed $\nu$-SVQR model.
         	\item[(c)] As the noise variance $\sigma$ increases, the RMSE obtained by $\nu$-SVQR model increases.
         \end{enumerate}
        
        \subsection*{Experiment 4}
        As stated in Remark-4, the proposed $\nu$-SVQR model is similar to the $\epsilon$-SVQR model in the sense that any solution $(\bar{w},\bar{b})$ obtained by the proposed $\nu$-SVQR can also be obtained by the $\epsilon$-SVQR. But, the proposed $\nu$-SVQR model has the capability of adjusting the $\epsilon$-insensitive zone according to the variance present in data. For realizing this direct benefit of proposed $\nu$-SVQR model over $\epsilon$-SVQR model, we perform the following experiment. 
                    
                    We generate 500 training data points of AD2 dataset where response points were polluted with noise from $U(-0.1,0.1)$. For predicting the $\tau=0.3$ quantile, we have tunned  parameters of $\epsilon$-QSVR as well as proposed $\nu$-SVQR model. We have found that at $\epsilon =0.1$ and $C=2^0$, the $\epsilon$-QSVR model obtains the minimum RMSE 0.0057. The proposed $\nu$-SVQR model obtains the RMSE value 0.0056 with parameters $\nu =0.5$ and $C=2^{0}*500$. The $\nu$-SVQR model obtains the asymmetric $\epsilon$ tube of width 0.0074. Now, with the same parameters setting in both  $\epsilon$-QSVR and $\nu$-SVQR model, we increase the variance present in noise of AD2 dataset to $U(-5,5)$. The $\epsilon$-SVQR model which has fixed $\epsilon$ value could obtain the RMSE 0.2168. But, the $\nu$-SVQR model automatically adjusts the width of asymmetric $\epsilon$-tube to 0.3734 and can obtain the RMSE value 0.1840.
                    
      \subsection*{Experiment 5}  The above experiments are enough to  empirically verify the claims made in this paper. But, we still want to check the performance of the  $\nu$-SVQR model on real world  data sets. For this, we have performed the experiments with the Servo (167$\times$5) dataset which is taken from UCI repository (Blake, \cite{UCIbenchmark}).  We have used 80$\%$ of  this dataset for training the proposed $\nu$-SVQR model and rest of the data points were used for the testing. The 100 random trails have been used to obtain the Error $E_{\tau}$ and Sparsity. Table (\ref{sevo_nusvqr}) shows the Error obtained by the proposed $\nu$-SVQR for different values of $\tau$ with different  value of $\nu$.  Table (\ref{servo_spars_nusvqr}) shows the sparsity obtained by the proposed $\nu$-SVQR model for different value of $\nu$ with  different $\tau$ values. It can be observed that irrespective of values of $\tau$, the sparsity decreases with increase in $\nu$ value .  
        
\section{Conclusions}          
We propose a novel $\nu$-Support Vector Quantile Regression ($\nu$-SVQR) model in this paper. There are several interesting properties of $\nu$-SVQR model which we have been briefly described and proved in this paper. Further, we have also empirically verified these proprieties by testing the proposed $\nu$-SVQR model on several artificial datasets as well as UCI dataset.
    \begin{landscape}

   	\begin{table}[]
   		\begin{adjustwidth}{-7cm}{}
   		\centering
   		 {\fontsize{9}{8} \selectfont
   		\begin{tabular}{|l|l|l|l|l|l|l|l|l|l|l|l|l|l|l|l|l|l|l|l|l|l|}
   			\hline
   	$\tau$	&	$\nu$                   & 0.050                   & 0.100 & 0.150 & 0.200 & 0.250 & 0.300 & 0.350 & 0.400 & 0.450 & 0.500 & 0.550 & 0.600 & 0.650 & 0.700 & 0.750 & 0.800 & 0.850 & 0.900 & 0.950 & 1.000        \\ \hline
   			\multirow{5}{*}{0.2} & $\epsilon$ & 0.384 & 0.288 & 0.225 & 0.190 & 0.156 & 0.143 & 0.120 & 0.108 & 0.091 & 0.076 & 0.068 & 0.055 & 0.047 & 0.043 & 0.037 & 0.027 & 0.019 & 0.013 & 0.003 & 0.000 \\ \cline{2-22} 
   			& SV                      & 0.080 & 0.125 & 0.175 & 0.230 & 0.290 & 0.335 & 0.380 & 0.425 & 0.480 & 0.525 & 0.580 & 0.630 & 0.685 & 0.725 & 0.770 & 0.840 & 0.885 & 0.925 & 0.980 & 1.060 \\ \cline{2-22} 
   			& Error                   & 0.030 & 0.075 & 0.130 & 0.185 & 0.230 & 0.275 & 0.330 & 0.370 & 0.425 & 0.470 & 0.520 & 0.570 & 0.620 & 0.660 & 0.710 & 0.775 & 0.820 & 0.860 & 0.920 & 0.940 \\ \cline{2-22} 
   			& RMSE                    & 0.160 & 0.072 & 0.050 & 0.038 & 0.044 & 0.052 & 0.063 & 0.063 & 0.060 & 0.058 & 0.059 & 0.058 & 0.060 & 0.061 & 0.063 & 0.064 & 0.065 & 0.066 & 0.067 & 0.067 \\ \cline{2-22} 
   			& MAE                     & 0.127 & 0.059 & 0.037 & 0.032 & 0.037 & 0.044 & 0.051 & 0.053 & 0.048 & 0.043 & 0.046 & 0.045 & 0.046 & 0.047 & 0.047 & 0.047 & 0.050 & 0.050 & 0.051 & 0.050 \\ \hline
   			\multirow{6}{*}{0.5} & $\nu$                      & 0.050 & 0.100 & 0.150 & 0.200 & 0.250 & 0.300 & 0.350 & 0.400 & 0.450 & 0.500 & 0.550 & 0.600 & 0.650 & 0.700 & 0.750 & 0.800 & 0.850 & 0.900 & 0.950 & 1.000 \\ \cline{2-22} 
   			& $\epsilon$ & 0.366 & 0.293 & 0.220 & 0.174 & 0.151 & 0.132 & 0.119 & 0.103 & 0.084 & 0.072 & 0.058 & 0.048 & 0.042 & 0.032 & 0.027 & 0.020 & 0.017 & 0.011 & 0.002 & 0.000 \\ \cline{2-22} 
   			& SV                      & 0.065 & 0.110 & 0.175 & 0.230 & 0.275 & 0.325 & 0.375 & 0.435 & 0.490 & 0.535 & 0.580 & 0.625 & 0.680 & 0.735 & 0.775 & 0.840 & 0.890 & 0.940 & 0.990 & 1.055 \\ \cline{2-22} 
   			& Error                   & 0.030 & 0.070 & 0.135 & 0.185 & 0.220 & 0.270 & 0.320 & 0.375 & 0.430 & 0.470 & 0.525 & 0.565 & 0.620 & 0.675 & 0.710 & 0.780 & 0.820 & 0.885 & 0.925 & 0.945 \\ \cline{2-22} 
   			& RMSE                    & 0.104 & 0.079 & 0.058 & 0.060 & 0.049 & 0.042 & 0.045 & 0.044 & 0.041 & 0.038 & 0.032 & 0.033 & 0.033 & 0.029 & 0.025 & 0.025 & 0.024 & 0.025 & 0.027 & 0.029 \\ \cline{2-22} 
   			& MAE                     & 0.090 & 0.063 & 0.049 & 0.046 & 0.035 & 0.032 & 0.036 & 0.035 & 0.029 & 0.028 & 0.023 & 0.026 & 0.025 & 0.023 & 0.020 & 0.020 & 0.019 & 0.020 & 0.022 & 0.024 \\ \hline
   			\multirow{6}{*}{0.7} & $\nu$                      & 0.050 & 0.100 & 0.150 & 0.200 & 0.250 & 0.300 & 0.350 & 0.400 & 0.450 & 0.500 & 0.550 & 0.600 & 0.650 & 0.700 & 0.750 & 0.800 & 0.850 & 0.900 & 0.950 & 1.000 \\ \cline{2-22} 
   			& $\epsilon$ & 0.363 & 0.281 & 0.243 & 0.196 & 0.164 & 0.147 & 0.126 & 0.107 & 0.090 & 0.079 & 0.067 & 0.052 & 0.045 & 0.036 & 0.029 & 0.025 & 0.016 & 0.012 & 0.005 & 0.000 \\ \cline{2-22} 
   			& SV                      & 0.080 & 0.120 & 0.175 & 0.225 & 0.285 & 0.330 & 0.380 & 0.430 & 0.485 & 0.530 & 0.590 & 0.630 & 0.680 & 0.730 & 0.790 & 0.835 & 0.890 & 0.925 & 0.985 & 1.045 \\ \cline{2-22} 
   			& Error                   & 0.035 & 0.085 & 0.130 & 0.185 & 0.235 & 0.275 & 0.320 & 0.375 & 0.420 & 0.470 & 0.525 & 0.570 & 0.615 & 0.665 & 0.720 & 0.765 & 0.830 & 0.860 & 0.920 & 0.955 \\ \cline{2-22} 
   			& RMSE                    & 0.128 & 0.090 & 0.091 & 0.090 & 0.075 & 0.063 & 0.057 & 0.047 & 0.044 & 0.042 & 0.043 & 0.047 & 0.046 & 0.045 & 0.043 & 0.043 & 0.044 & 0.046 & 0.047 & 0.046 \\ \cline{2-22} 
   			& MAE                     & 0.103 & 0.070 & 0.070 & 0.069 & 0.053 & 0.048 & 0.043 & 0.037 & 0.033 & 0.032 & 0.033 & 0.033 & 0.031 & 0.031 & 0.029 & 0.030 & 0.030 & 0.031 & 0.033 & 0.033 \\ \hline
   			\multirow{6}{*}{0.8} & $\nu$                      & 0.050 & 0.100 & 0.150 & 0.200 & 0.250 & 0.300 & 0.350 & 0.400 & 0.450 & 0.500 & 0.550 & 0.600 & 0.650 & 0.700 & 0.750 & 0.800 & 0.850 & 0.900 & 0.950 & 1.000 \\ \cline{2-22} 
   			& $\epsilon$ & 0.367 & 0.309 & 0.244 & 0.214 & 0.190 & 0.161 & 0.145 & 0.119 & 0.096 & 0.084 & 0.075 & 0.056 & 0.049 & 0.037 & 0.031 & 0.026 & 0.014 & 0.012 & 0.004 & 0.000 \\ \cline{2-22} 
   			& SV                      & 0.080 & 0.130 & 0.175 & 0.220 & 0.280 & 0.335 & 0.380 & 0.435 & 0.485 & 0.530 & 0.570 & 0.635 & 0.695 & 0.730 & 0.780 & 0.835 & 0.890 & 0.935 & 0.990 & 1.060 \\ \cline{2-22} 
   			& Error                   & 0.035 & 0.090 & 0.135 & 0.175 & 0.230 & 0.280 & 0.325 & 0.385 & 0.430 & 0.470 & 0.515 & 0.575 & 0.625 & 0.670 & 0.715 & 0.775 & 0.820 & 0.870 & 0.920 & 0.940 \\ \cline{2-22} 
   			& RMSE                    & 0.143 & 0.131 & 0.082 & 0.087 & 0.083 & 0.074 & 0.071 & 0.064 & 0.061 & 0.061 & 0.061 & 0.051 & 0.055 & 0.054 & 0.053 & 0.054 & 0.057 & 0.058 & 0.058 & 0.058 \\ \cline{2-22} 
   			& MAE                     & 0.112 & 0.102 & 0.060 & 0.062 & 0.060 & 0.058 & 0.053 & 0.046 & 0.043 & 0.044 & 0.044 & 0.039 & 0.042 & 0.041 & 0.040 & 0.042 & 0.044 & 0.044 & 0.044 & 0.044 \\ \hline
   		\end{tabular}}
   	\caption{Performance of the proposed $\nu$-QSVR model on AD1 dataset for different $\tau$ values.}
   	\label{table1}
 \end{adjustwidth}
   	\end{table}
   \end{landscape}

	\begin{table}[]
		\centering
		{\fontsize{10}{10} \selectfont
		\begin{tabular}{|l|l|l|l|l|l|l|l|}
			\hline
		$\tau$                  &  \textit{l}       & 100   & 200   & 500  & 1000 & 3000 & 5000 \\ \hline
			\multirow{5}{*}{0.1} & SV      & 0.85  & 0.83  & 0.81 & 0.81 & 0.80 & 0.80 \\ \cline{2-8} 
			& Error   & 0.75  & 0.78  & 0.79 & 0.79 & 0.80 & 0.80 \\ \cline{2-8} 
			& Ratio   & 14.00 & 13.09 & 9.97 & 9.72 & 9.20 & 9.17 \\ \cline{2-8} 
			& $\epsilon$ & 0.04  & 0.05  & 0.05 & 0.04 & 0.05 & 0.05 \\ \cline{2-8} 
			& RMSE    & 0.11  & 0.05  & 0.06 & 0.05 & 0.01 & 0.01 \\ \hline
			\multirow{6}{*}{0.3} &\textit{l}        & 100   & 200   & 500  & 1000 & 3000 & 5000 \\ \cline{2-8} 
			& SV      & 0.85  & 0.82  & 0.82 & 0.81 & 0.80 & 0.80 \\ \cline{2-8} 
			& Error   & 0.73  & 0.77  & 0.79 & 0.80 & 0.80 & 0.80 \\ \cline{2-8} 
			& Ratio   & 2.65  & 2.56  & 2.41 & 2.38 & 2.36 & 2.35 \\ \cline{2-8} 
			& $\epsilon$ & 0.04  & 0.03  & 0.03 & 0.03 & 0.03 & 0.03 \\ \cline{2-8} 
			& RMSE    & 0.10  & 0.05  & 0.03 & 0.04 & 0.01 & 0.01 \\ \hline
			\multirow{6}{*}{0.7} & \textit{l}       & 100   & 200   & 500  & 1000 & 3000 & 5000 \\ \cline{2-8} 
			& SV      & 0.85  & 0.82  & 0.82 & 0.81 & 0.80 & 0.80 \\ \cline{2-8} 
			& Error   & 0.73  & 0.76  & 0.79 & 0.79 & 0.80 & 0.80 \\ \cline{2-8} 
			& Ratio   & 0.35  & 0.38  & 0.42 & 0.43 & 0.43 & 0.43 \\ \cline{2-8} 
			& $\epsilon$ & 0.02  & 0.03  & 0.03 & 0.03 & 0.03 & 0.03 \\ \cline{2-8} 
			& RMSE    & 0.07  & 0.05  & 0.04 & 0.02 & 0.01 & 0.01 \\ \hline
			\multirow{6}{*}{0.9} & \textit{l}        & 100   & 200   & 500  & 1000 & 3000 & 5000 \\ \cline{2-8} 
			& SV      & 0.87  & 0.82  & 0.81 & 0.81 & 0.80 & 0.80 \\ \cline{2-8} 
			& Error   & 0.76  & 0.77  & 0.79 & 0.79 & 0.80 & 0.80 \\ \cline{2-8} 
			& Ratio   & 0.09  & 0.08  & 0.10 & 0.11 & 0.11 & 0.11 \\ \cline{2-8} 
			& $\epsilon$ & 0.03  & 0.04  & 0.04 & 0.05 & 0.05 & 0.05 \\ \cline{2-8} 
			& RMSE    & 0.13  & 0.06  & 0.07 & 0.03 & 0.02 & 0.02 \\ \hline
		\end{tabular}}
	\caption{Performance of the proposed $\nu$-SVQR with different size of training set.}
	\label{table2}
	\end{table}
	\begin{table}[]
		\centering
			{\fontsize{10}{10} \selectfont
		\begin{tabular}{|l|l|l|l|l|l|l|l|l|l|l|l|}
			\hline
			$\tau$                 & $\sigma$   & 0.1  & 0.2  & 0.3  & 0.4  & 0.5  & 0.6  & 0.7  & 0.8  & 0.9  & 1    \\ \hline
			\multirow{4}{*}{0.9} & $\epsilon$ & 0.02 & 0.04 & 0.05 & 0.07 & 0.09 & 0.11 & 0.12 & 0.14 & 0.16 & 0.18 \\ \cline{2-12} 
			& Error   & 0.39 & 0.39 & 0.40 & 0.39 & 0.39 & 0.39 & 0.39 & 0.39 & 0.39 & 0.39 \\ \cline{2-12} 
			& SV      & 0.41 & 0.41 & 0.42 & 0.41 & 0.41 & 0.41 & 0.41 & 0.41 & 0.41 & 0.42 \\ \cline{2-12} 
			& RMSE    & 0.01 & 0.02 & 0.02 & 0.03 & 0.04 & 0.05 & 0.05 & 0.06 & 0.07 & 0.08 \\ \hline
			\multirow{5}{*}{0.7} & $\sigma$   & 0.1  & 0.2  & 0.3  & 0.4  & 0.5  & 0.6  & 0.7  & 0.8  & 0.9  & 1    \\ \hline
			& $\epsilon$ & 0.01 & 0 .03 & 0.04 & 0.06 & 0.07 & 0.09 & 0.10 & 0.12 & 0.13 & 0.15 \\ \cline{2-12} 
			& Error   & 0.39 & 0.39 & 0.39 & 0.39 & 0.39 & 0.39 & 0.39 & 0.39 & 0.39 & 0.39 \\ \cline{2-12} 
			& SV      & 0.41 & 0.41 & 0.41 & 0.41 & 0.41 & 0.41 & 0.41 & 0.41 & 0.41 & 0.41 \\ \cline{2-12} 
			& RMSE    & 0.00 & 0.01 & 0.02 & 0.02 & 0.03 & 0.03 & 0.04 & 0.04 & 0.05 & 0.05 \\ \hline
			\multirow{5}{*}{0.5} & $\sigma$   & 0.1  & 0.2  & 0.3  & 0.4  & 0.5  & 0.6  & 0.7  & 0.8  & 0.9  & 1    \\  \hline 
			& $\epsilon$ & 0.01 & 0.03 & 0.05 & 0.06 & 0.07 & 0.09 & 0.10 & 0.12 & 0.13 & 0.15 \\ \cline{2-12} 
			& Error   & 0.39 & 0.39 & 0.39 & 0.39 & 0.39 & 0.39 & 0.39 & 0.39 & 0.39 & 0.39 \\ \cline{2-12} 
			& SV      & 0.41 & 0.42 & 0.42 & 0.41 & 0.41 & 0.41 & 0.41 & 0.41 & 0.41 & 0.41 \\ \cline{2-12} 
			& RMSE    & 0.00 & 0.01 & 0.01 & 0.02 & 0.02 & 0.03 & 0.03 & 0.03 & 0.04 & 0.04 \\ \hline
			\multirow{6}{*}{0.3} & $\sigma$   & 0.1  & 0.2  & 0.3  & 0.4  & 0.5  & 0.6  & 0.7  & 0.8  & 0.9  & 1    \\  \hline
			& $\epsilon$ & 0.01 & 0.03 & 0.05 & 0.07 & 0.08 & 0.10 & 0.11 & 0.13 & 0.15 & 0.16 \\ \cline{2-12} 
			& Error   & 0.39 & 0.39 & 0.39 & 0.39 & 0.39 & 0.39 & 0.39 & 0.39 & 0.39 & 0.39 \\ \cline{2-12} 
			& SV      & 0.42 & 0.41 & 0.41 & 0.41 & 0.41 & 0.41 & 0.41 & 0.41 & 0.41 & 0.41 \\ \cline{2-12} 
			& RMSE    & 0.01 & 0.01 & 0.02 & 0.02 & 0.03 & 0.03 & 0.04 & 0.04 & 0.05 & 0.06 \\ \hline 
			& $\sigma$   & 0.1  & 0.2  & 0.3  & 0.4  & 0.5  & 0.6  & 0.7  & 0.8  & 0.9  & 1    \\ \hline  
			\multirow{4}{*}{0.1} & $\epsilon$ & 0.02 & 0.04 & 0.06 & 0.08 & 0.10 & 0.12 & 0.14 & 0.16 & 0.18 & 0.20 \\ \cline{2-12} 
			& Error   & 0.39 & 0.39 & 0.39 & 0.39 & 0.39 & 0.39 & 0.39 & 0.39 & 0.39 & 0.39 \\ \cline{2-12} 
			& SV      & 0.41 & 0.41 & 0.41 & 0.41 & 0.41 & 0.41 & 0.41 & 0.41 & 0.41 & 0.41 \\ \cline{2-12} 
			& RMSE    & 0.01 & 0.02 & 0.03 & 0.04 & 0.05 & 0.06 & 0.07 & 0.08 & 0.09 & 0.10 \\ \hline
		\end{tabular}}
	\caption{Performance of the proposed $\nu$-SVQR with different values of noise variance $\sigma$.}
	\label{table3}
	\end{table}


\begin{figure}
	\centering
	\begin{tabular}{cc}
		\subfloat[] {\includegraphics[width = 5.0in,height=1.0in]{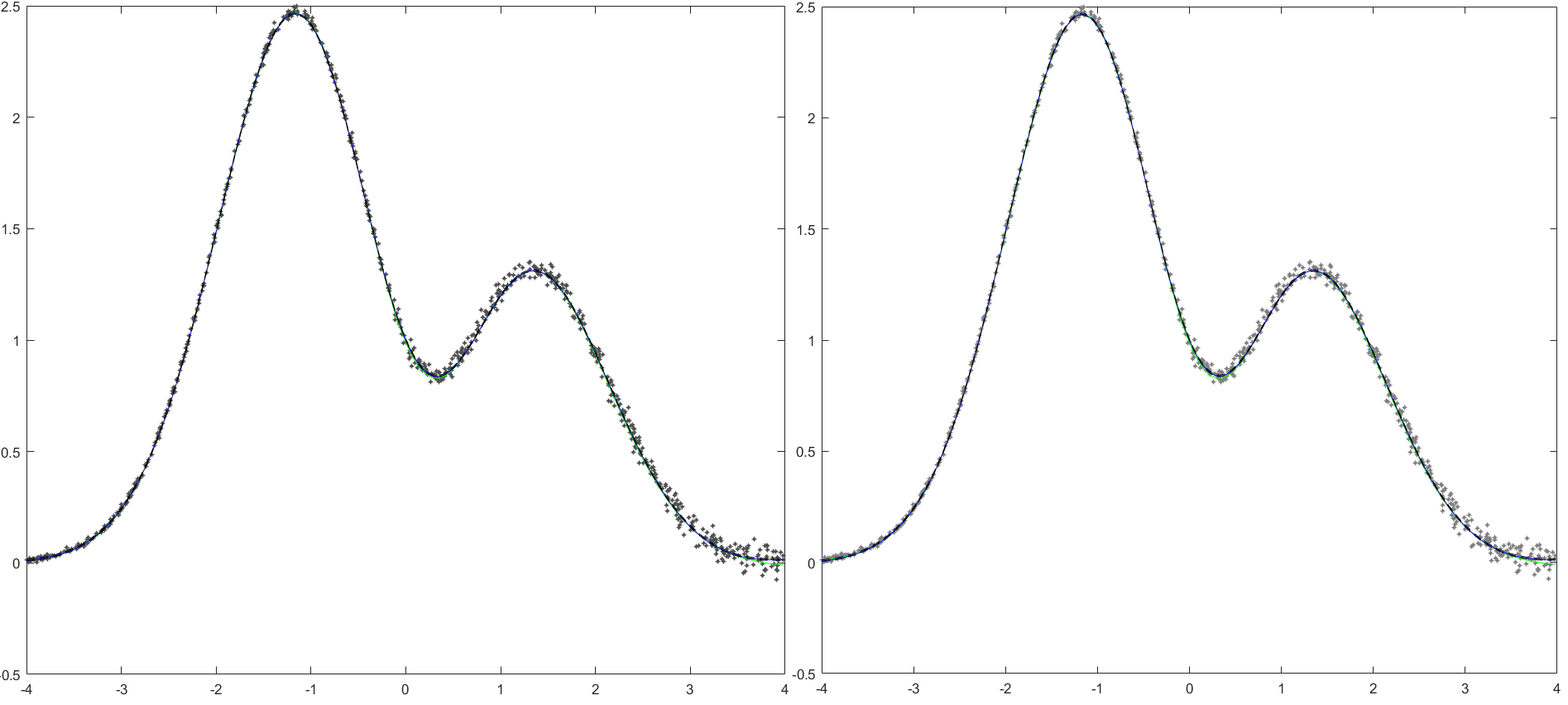}}\\
		\subfloat[] {\includegraphics[width = 5.0in,height=4.5in]{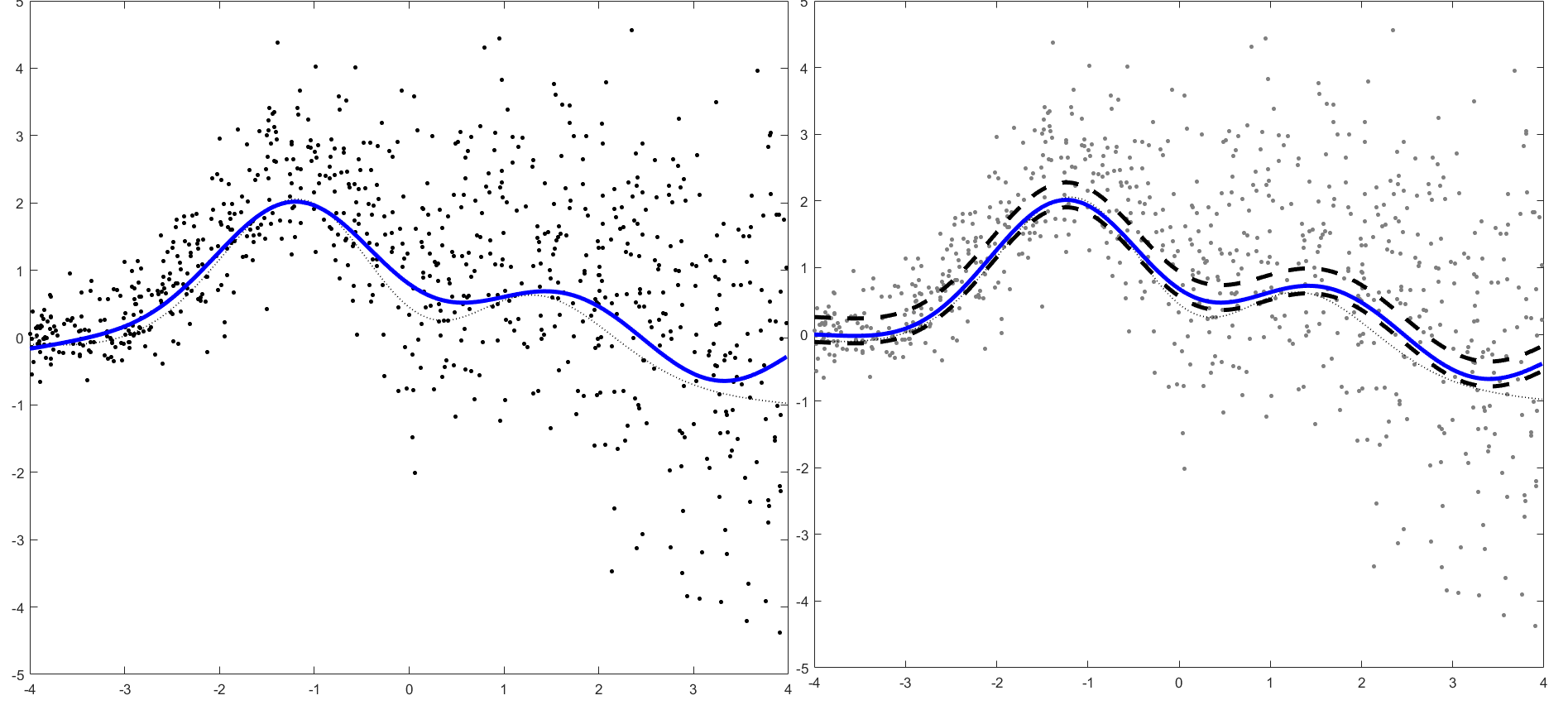}}
	\end{tabular}
	\caption{The $\epsilon$-SVQR model (left) fails to adjust the total width of the asymmetric $\epsilon$-insensitive tube with increase in noise variance. The proposed $\epsilon$-SVQR model (right) adjusts the total width of the asymmetric $\epsilon$-insensitive tube according to increase in noise variance and hence can obtain better estimate.}
\end{figure}


	\begin{table}[]
		\centering
		\begin{tabular}{|l|l|l|l|l|l|l|l|l|l|}
			\hline
			$\nu$/$\tau$ & 0.1   & 0.2   & 0.3   & 0.4   & 0.5   & 0.6   & 0.7   & 0.8   & 0.9   \\ \hline
			0.10    & 0.174 & 0.445 & 0.464 & 0.401 & 0.316 & 0.222 & 0.129 & 0.056 & 0.054 \\ \hline
			0.15   & 0.169 & 0.403 & 0.452 & 0.399 & 0.316 & 0.221 & 0.129 & 0.056 & 0.055 \\ \hline
			0.20    & 0.040 & 0.248 & 0.395 & 0.387 & 0.315 & 0.220 & 0.129 & 0.056 & 0.057 \\ \hline
			0.25   & 0.037 & 0.062 & 0.187 & 0.313 & 0.300 & 0.217 & 0.129 & 0.056 & 0.059 \\ \hline
			0.30    & 0.040 & 0.073 & 0.087 & 0.171 & 0.251 & 0.210 & 0.129 & 0.056 & 0.059 \\ \hline
			0.35   & 0.043 & 0.072 & 0.066 & 0.092 & 0.160 & 0.195 & 0.127 & 0.056 & 0.060 \\ \hline
			0.40    & 0.043 & 0.069 & 0.069 & 0.069 & 0.094 & 0.161 & 0.123 & 0.056 & 0.061 \\ \hline
			0.45   & 0.044 & 0.059 & 0.065 & 0.067 & 0.076 & 0.112 & 0.119 & 0.056 & 0.062 \\ \hline
			0.50    & 0.045 & 0.055 & 0.062 & 0.069 & 0.071 & 0.076 & 0.116 & 0.057 & 0.062 \\ \hline
			0.55   & 0.043 & 0.062 & 0.062 & 0.067 & 0.081 & 0.066 & 0.095 & 0.058 & 0.064 \\ \hline
			0.60    & 0.038 & 0.066 & 0.062 & 0.068 & 0.083 & 0.070 & 0.072 & 0.060 & 0.067 \\ \hline
			0.65   & 0.038 & 0.058 & 0.059 & 0.067 & 0.081 & 0.085 & 0.074 & 0.061 & 0.067 \\ \hline
			0.70    & 0.038 & 0.057 & 0.060 & 0.069 & 0.082 & 0.092 & 0.079 & 0.065 & 0.070 \\ \hline
			0.75   & 0.038 & 0.054 & 0.061 & 0.070 & 0.082 & 0.094 & 0.081 & 0.072 & 0.070 \\ \hline
			0.80    & 0.038 & 0.054 & 0.061 & 0.074 & 0.082 & 0.092 & 0.082 & 0.072 & 0.073 \\ \hline
			0.85   & 0.038 & 0.055 & 0.064 & 0.071 & 0.081 & 0.091 & 0.081 & 0.067 & 0.075 \\ \hline
			0.90    & 0.041 & 0.054 & 0.064 & 0.068 & 0.084 & 0.085 & 0.078 & 0.068 & 0.075 \\ \hline
			0.95   & 0.040 & 0.055 & 0.065 & 0.071 & 0.082 & 0.083 & 0.072 & 0.068 & 0.074 \\ \hline
			1.00      & 0.040 & 0.055 & 0.067 & 0.071 & 0.078 & 0.084 & 0.073 & 0.069 & 0.073 \\ \hline
		\end{tabular}
	\caption{Performance of the proposed $\nu$-SVQR model with different value of $\nu$ for different $\tau$ values.}
	\label{sevo_nusvqr}
	\end{table}


	\begin{table}[]
		\centering
		\begin{tabular}{|l|l|l|l|l|l|l|l|l|l|}
			\hline
			$\nu$/$\tau$  & 0.1   & 0.2   & 0.3   & 0.4   & 0.5   & 0.6   & 0.7   & 0.8   & 0.9   \\ \hline
			0.1    & 89.47 & 88.72 & 89.47 & 89.47 & 89.47 & 89.47 & 88.72 & 88.72 & 89.47 \\ \hline
			0.15   & 83.46 & 82.71 & 84.21 & 83.46 & 83.46 & 83.46 & 84.21 & 84.21 & 84.21 \\ \hline
			0.2    & 78.95 & 78.95 & 79.70 & 78.20 & 78.95 & 78.95 & 79.70 & 78.95 & 78.95 \\ \hline
			0.25   & 74.44 & 74.44 & 73.68 & 74.44 & 73.68 & 73.68 & 72.93 & 73.68 & 74.44 \\ \hline
			0.3    & 69.17 & 68.42 & 69.17 & 69.17 & 69.17 & 69.17 & 69.17 & 69.17 & 69.17 \\ \hline
			0.35   & 63.91 & 63.91 & 63.91 & 63.91 & 63.91 & 63.91 & 63.16 & 63.91 & 64.66 \\ \hline
			0.4    & 57.89 & 57.89 & 57.89 & 57.89 & 58.65 & 58.65 & 59.40 & 59.40 & 58.65 \\ \hline
			0.45   & 54.14 & 52.63 & 52.63 & 53.38 & 52.63 & 54.14 & 54.14 & 54.14 & 54.14 \\ \hline
			0.5    & 48.87 & 48.12 & 48.87 & 48.12 & 47.37 & 48.12 & 49.62 & 48.87 & 48.12 \\ \hline
			0.55   & 42.86 & 42.86 & 43.61 & 43.61 & 43.61 & 44.36 & 42.86 & 43.61 & 43.61 \\ \hline
			0.6    & 38.35 & 38.35 & 37.59 & 36.84 & 38.35 & 38.35 & 38.35 & 39.10 & 39.10 \\ \hline
			0.65   & 33.08 & 33.08 & 33.83 & 32.33 & 33.08 & 32.33 & 33.08 & 34.59 & 34.59 \\ \hline
			0.7    & 27.82 & 29.32 & 27.07 & 27.82 & 27.07 & 27.82 & 27.82 & 29.32 & 29.32 \\ \hline
			0.75   & 22.56 & 22.56 & 21.80 & 22.56 & 23.31 & 22.56 & 23.31 & 23.31 & 24.06 \\ \hline
			0.8    & 17.29 & 17.29 & 17.29 & 18.80 & 17.29 & 16.54 & 18.05 & 18.80 & 18.80 \\ \hline
			0.85   & 13.53 & 12.03 & 12.03 & 11.28 & 13.53 & 12.78 & 12.03 & 13.53 & 14.29 \\ \hline
			0.9    & 8.27  & 9.02  & 8.27  & 8.27  & 7.52  & 7.52  & 7.52  & 8.27  & 9.02  \\ \hline
			0.95   & 3.01  & 2.26  & 3.76  & 3.01  & 3.01  & 3.01  & 3.01  & 3.76  & 4.51  \\ \hline
			1      & 0.00  & 0.00  & 0.00  & 0.00  & 0.00  & 0.00  & 0.00  & 0.00  & 0.00  \\ \hline
		\end{tabular}
		\caption{Sparsity obtained by the proposed $\nu$-SVQR model with different value of $\nu$ for different $\tau$ values.}
		\label{servo_spars_nusvqr}
	\end{table}

     \section*{Acknowledgment}
          We would like to acknowledge Ministry of Electronics and Information Technology, Government of India, as this work has been funded by them under Visvesvaraya PhD Scheme for Electronics and IT, Order No. Phd-MLA/4(42)/2015-16. 
      \section* {Conflict of Interest}
       We authors hereby declare that we do not have any conflict of interest with the content of this manuscript.

\end{document}